# Predicting Mental Health Help-Seeking Behaviour on Reddit: A Survival Analysis of Linguistic and Behavioural Signals

Nisar Ahmed[a,*], Ammara Kazmi[b], Irtaza Bashir Raja[c], Didier Devaurs[a]
[a] Department of Computer and Information Sciences, University of Strathclyde, Glasgow, UK
[b] Department of Computer Science, Sir Syed CASE Institute of Technology, Islamabad, Pakistan
[c] Department of Engineering Management, NUST, Rawalpindi, Pakistan
[*] Corresponding author: ch.nisar89@icloud.com

## Abstract

**Background and Objective:** Most prior work on predicting mental health help-seeking treats it as post-hoc classification on users already known to be in clinical communities. A design that cannot answer whether someone approaching a transition can be identified in advance is limited in scope. Instead, we frame the problem as a time-to-event task and ask two questions: which Reddit users will eventually seek mental health support, and how imminent is that transition?

**Methods:** We construct a two-arm cohort from Reddit: a control frame of 6,337 ordinary users with genuine right-censoring (12.1% transition rate) to answer who will transition, and a clinical frame of 5,023 eventual transitioners to answer how soon. Features are restricted strictly to posts written before any clinical activity; user-grouped five-fold cross-validation with a four-year temporal holdout prevents data leakage. We compare term frequency–inverse document frequency (TF-IDF) bag-of-words features, MentalBERT embeddings (a mental-health-domain-adapted variant of Bidirectional Encoder Representations from Transformers, BERT), and behavioural activity features using Cox proportional hazards models, Random Survival Forests, and a long short-term memory (LSTM) sequential model.

**Results**: Behavioural features alone outperform text features for predicting who will eventually seek help (C-index 0.797±0.018 vs. 0.763), and combining both is best of all (0.823±0.018 under cross-validation, rising to 0.856 on a four-year future-data holdout). The pattern reverses for predicting how soon a known transition will happen: among clinical-frame users, text remains a stable predictor (C-index 0.698±0.003) while behavioural features become unstable and uninformative.

**Conclusions:** Structural engagement patterns carry predictive information about future help-seeking beyond linguistic content. Separating *who* will seek help from *when* yields a cleaner analysis, and the combined model's stability under temporal holdout suggests time-invariant behavioural signals underpin the primary result.

## 1. Introduction

Mental health conditions affect approximately one in eight people globally [1], yet delays between symptom onset and first professional contact are frequently measured in years rather than months across conditions: recent meta-analyses report a median 3.5-year delay to help-seeking and 6.7 years to diagnosis in bipolar disorder (n>40,000) [2], and a mean 6.7-year duration of untreated illness in obsessive-compulsive disorder (n=2,750) [3]. Social media platforms, particularly Reddit, have emerged as spaces where individuals discuss psychological distress in natural language before, during, and after any formal help-seeking [4]. This raises a compelling question for computational researchers: can naturally occurring online language and behaviour predict an individual's transition into a mental health community?

Prior work on this problem has largely proceeded as *post-hoc classification*: given a user known to post in a clinical subreddit, classify whether a given post belongs to a clinical category [5]. This framing conflates discovery and prediction, samples without a denominator, and cannot answer whether a user who has *not yet* sought help is approaching a transition. Chancellor and De Choudhury [6] review the field and note these systematic limitations: no control groups, no temporal prospectivity, and outcome labels that proxy clinical states rather than measuring them.

Here, we address these limitations with a survival analysis framework. We define help-seeking as the event of *first posting to a clinical community*; users who have never done so by the cutoff date are right-censored. This framing naturally accommodates the time-to-event structure of the problem and allows calibrated probability statements about future transitions.

This study addresses two research questions. First, can pre-transition linguistic and behavioural features in non-clinical posts predict which Reddit users will eventually post in a clinical community, and how accurately can they be ranked by transition probability (RQ1, who?). Second, for users who are known to eventually transition, can the same features predict how imminent that transition is (RQ2, how soon?). RQ1 is answered with a control-frame analysis (Arm 1) that provides an unbiased population-level estimate of transition incidence, with genuine right-censoring; RQ2 is answered with a clinical-frame analysis (Arm 2) on the subset of users who transitioned, exploiting the rich pre-transition text to estimate time remaining.

The contributions of this paper are as follows:

1. A two-arm cohort design that decouples incidence estimation from imminence estimation and provides a proper denominator for survival analysis.
2. An empirical demonstration that behavioural features, such as activity patterns and community diversity (number of unique subreddits visited), outperform text features for survival ranking (C-index 0.797 vs. 0.763), with a sensitivity analysis confirming robustness to a potentially confounded recency feature.
3. A combined text-and-behaviour model achieving C-index 0.823 under cross-validation (CV) and 0.856 on a four-year temporal holdout.
4. A stable imminence model (Arm 2, C-index 0.698) showing that text content, not behavioural patterns, carries the timing signal.

Beyond its methodological contribution, this framing has practical relevance: a model that reliably ranks who is likely to seek help, and how soon, could inform lightweight, privacy-preserving outreach on social platforms, for example prioritising peer-support resources toward higher-risk communities rather than individuals, without any individual-level clinical claim. The same two-arm survival design is not specific to Reddit or to mental health; it applies wherever a platform can observe a leakage-safe pre-event history and a well-defined transition event, including clinical time-to-event settings such as treatment-free remission prediction in chronic disease.

## 2. Related work

### 2.1. Mental health on social media

Park et al. [4] found that depressed and non-depressed Twitter users perceive and use the platform differently, with depressed users favouring emotional and social-support interaction over the information-sharing use favoured by non-depressed users. De Choudhury and De [7] examined self-disclosure and social support dynamics on Reddit specifically. De Choudhury et al. [8] demonstrated that linguistic cues on Twitter correlate with clinical depression assessments, establishing the basic feasibility of social-media-based mental health research. Guntuku et al. [9] reviewed over 200 studies linking social media signals to mental health outcomes and identified selection bias and lack of longitudinal design as the primary threats to validity.

### 2.2. NLP for mental health prediction

Coppersmith et al. [5] quantified mental health signals in Twitter data, framing depression and post-traumatic stress disorder (PTSD) detection as a text classification problem and laying the groundwork for the Computational Linguistics and Clinical Psychology (CLPsych) shared task series that followed; a later iteration targeted suicide risk specifically [10]. Tlachac and Rundensteiner [11] developed retrospective text-based screening instruments for depression. Owen et al. [12] evaluated transformer-based language models for longitudinal depression detection in web-based forums. Naseem et al. [13] proposed ordinal classification for early identification of depression severity levels on Reddit.

Common limitations across this literature are the absence of a proper control group, reliance on proxy labels, and post-hoc rather than prospective framing. Chancellor and De Choudhury [6] synthesised these concerns and called for ecologically valid, longitudinal designs. Our study is a direct response to their call.

### 2.3. Survival analysis for clinical prediction

Cox proportional hazards regression [14] is the standard statistical model for time-to-event prediction in medicine [15]. Harrell et al. [16] introduced the concordance index (C-index) as an appropriate discrimination metric to assess time-to-event prediction models. Graf et al. [17] proposed the Brier score for calibration of survival models. Survival models are standard in clinical prediction (e.g., for disease progression [18]), but their application to social media mental health data is rare. One exception is Liu et al. [19], who fitted a Cox model to Reddit posting sequences preceding transitions to r/SuicideWatch and reported a concordance index of 0.5123 only, which is close to random chance. As their design is restricted to users who eventually post to r/SuicideWatch, it *lacks a genuine non-transitioning control population* and does not separate incidence from imminence.

### 2.4. Domain-adapted language models

Ji et al. [20] trained MentalBERT and MentalRoBERTa on mental health text from Reddit, demonstrating that domain-adapted pretraining improves performance on clinical natural language processing (NLP) benchmarks over the generic BERT. Here, we use MentalBERT as our embedding model rather than the generic BERT, motivated by the domain specificity of our task.

### 2.5. Research gap and study positioning

Across this literature, *three limitations reoccur*. *First*, studies sample users exclusively from clinical communities, producing cohorts with no genuine controls and no population-level denominator, making it impossible to estimate how many ordinary users will ever transition. *Second*, the problem is cast as classification rather than time-to-event prediction, discarding the temporal structure that is clinically most informative. *Third*, who will eventually seek help and how soon are treated as a single question rather than two distinct ones that may be answered by different signals. Our study addresses all three limitations, as it involves a two-arm cohort design with proper censoring that separates incidence estimation from imminence prediction, and a survival analysis that preserves the time dimension throughout.

## 3. Data

This study uses a two-arm survival design: control-frame users are drawn from neutral subreddits, and clinical-frame users from clinical subreddits, each with a pre-index posting history preceding an index time. Arm 1 asks which control-frame users will eventually seek help; Arm 2 asks how soon, among clinical-frame users who have already sought help. Fig. 1 previews the full pipeline, including the feature extraction and modelling steps described in Section 4.

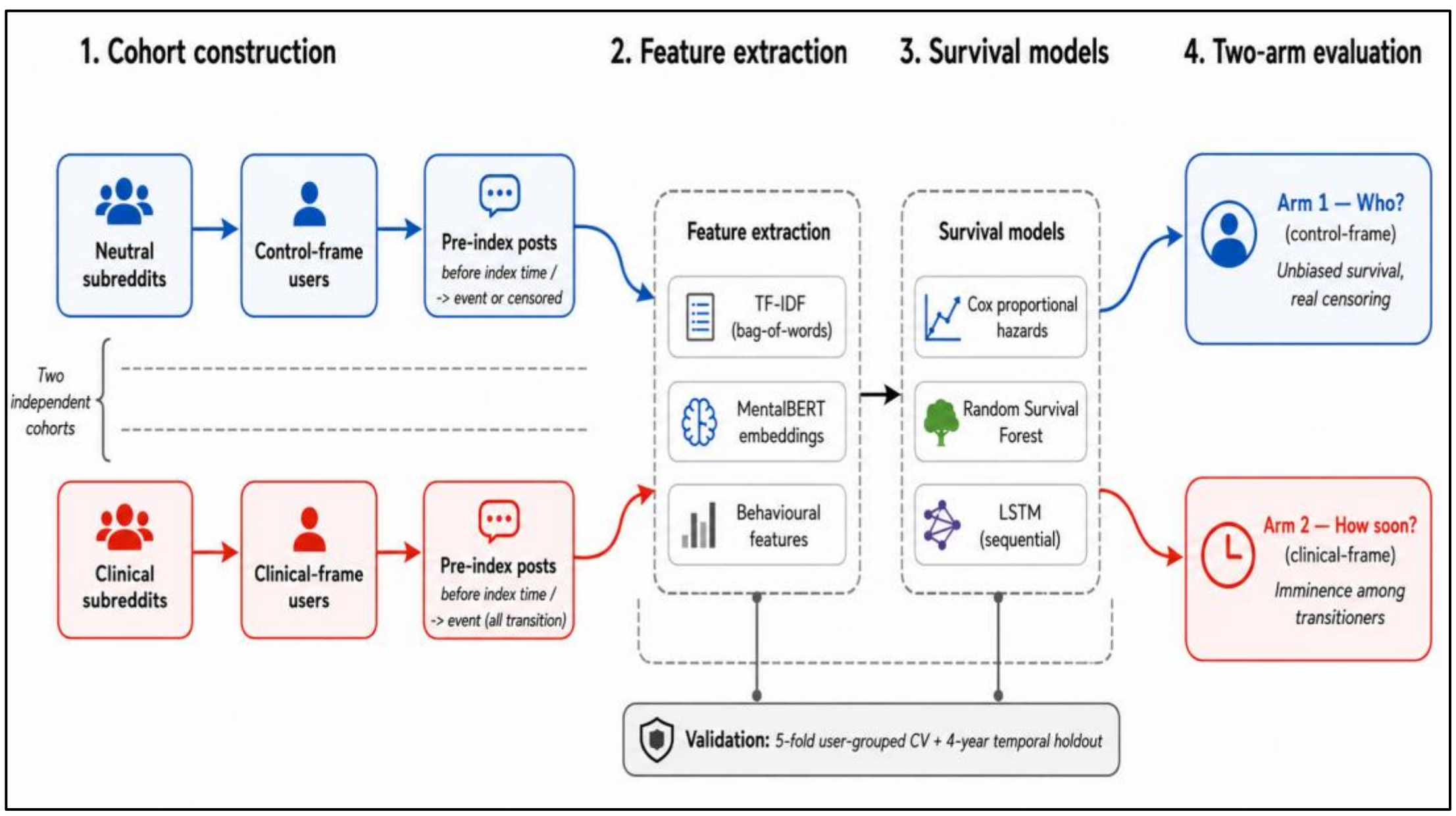

**Fig. 1.** Overview of the two-arm study pipeline, from cohort construction through feature extraction, modelling, and evaluation.

### 3.1. Collection and ethical considerations

We collected Reddit data via the PRAW (Python Reddit API Wrapper) API in June 2026, reading only publicly accessible posts. No private information or user identifiers beyond pseudonymous usernames were collected, and no contact attempts were made. Data are stored locally and used solely for aggregate statistical analysis; no individual-level clinical inferences are made. No ethical approval was sought, as the study involved no interaction with individuals (see Section 7.4). We discuss dual-use considerations in Section 7.

We define two *discovery frames*:

The clinical frame involves users discovered by sampling recent posts from ten clinical subreddits: r/depression, r/Anxiety, r/bipolar, r/ADHD, r/BipolarReddit, r/OCD, r/PTSD, r/socialanxiety, r/mentalhealth, and r/SuicideWatch. Posts from r/SuicideWatch are used only as label signals and are never displayed, quoted, or analysed individually.

The control frame involves users discovered by sampling recent posts from ten neutral communities: r/AskReddit, r/CasualConversation, r/NoStupidQuestions, r/todayilearned, r/mildlyinteresting, r/explainlikeimfive, r/books, r/movies, r/Cooking, and r/travel. These communities attract broad populations discussing everyday experiences without a clinical focus.

For each discovered user, we retrieved their full publicly available history of both submissions and comments (8,579,850 items in total: 1,509,610 submissions and 7,070,240 comments); throughout, "post" refers to either. A user's *category* is assigned on a per post basis: any post to a clinical subreddit is labelled CLINICAL; all others are labelled NON-CLINICAL. A small number of users ($n = 38$) were independently sampled from both frames `(Discovered_Via = both)`; they are counted separately in Table 1 and excluded from single-arm analyses.

For each post, we record the following fields: `User`, `Title`, `Body`, `Created_UTC (Unix epoch integer)`, `Subreddit`, `Category`, `Discovered_Via`, `Type (submission/comment)`, `Id`, `Score (upvote count)`. When Body is null (e.g., in link posts), Title is used as the sole content; such posts are retained rather than dropped, as their count and metadata still carry behavioural signal.

## 3.2. Cohort construction and survival framing

For each user $u$ we define:

$$t_{\text{first-clinical}}(u) = \min\{\text{Created_UTC}(p) : \text{Category}(p) = \text{Clinical},\ p \in \text{posts}(u)\}$$

$$\text{event}(u) = \mathbf{1}[t_{\text{first-clinical}}(u) \text{ exists}]$$

The *index time* is the event time for cases of users seeking help, or the cutoff date ($t_{\text{cutoff}} =$ June 2026) for censored users. *Lead time* (duration) is measured from the user's first ever observed post to the index time.

**Critical invariant:** all features are computed exclusively from posts with `Created_UTC` < `index_time`. No post-index information enters any feature vector. This prevents temporal leakage and is enforced by the pre-index post extraction stage.

**Modellable subset.** Users with no pre-index posts (i.e., for whom the first observed post *was* the clinical post) carry no text signal and are excluded. Zero-duration cases (i.e., index time ≤ first post time, arising from API timestamp rounding) are also excluded. Applying these filters to the 6,109 clinical-frame users yields 5,023 modellable Arm 2 users; for Arm 1, applying the same filters to the 6,362 control-frame users yields 6,337 modellable users (766 events, 12.1% event rate).

## 3.3. Descriptive statistics

This section reports cohort-level descriptive statistics characterising the two discovery frames before turning to survival modelling.

Table 1 shows the cohort composition. The control frame provides 6,362 users with a 12.4% transition rate; a realistic population-level incidence that enables unbiased survival estimation. Clinical-frame users are enriched for cases (96.2%) and have much shorter lead times (median 335 days vs. 577 days for control-frame users), reflecting that they were recruited near the time of their clinical activity. The median lead time of 577 days (interquartile range (IQR) 174–1342) in the control frame indicates that transitions can be preceded by more than a year of non-clinical posting, leaving substantial room for prospective risk stratification in principle (though the retrospective nature of this study means we cannot make direct early-detection claims).

**Table 1.** Cohort-wide descriptive statistics for each discovery frame. Med. Posts is the median number of pre-index posts per user within that frame. Median lead time and associated interquartile range (IQR) are reported in days for cases only. Average post length is in characters (Title + Body).

| Frame | *Users* | Events | Rate | Med. posts | Avg. chars | Med. lead (d) | IQR (d) |
|---|---|---|---|---|---|---|---|
| Control | 6,362 | 791 | 12.4% | 36 | 194 | 577 | 174–1342 |
| Clinical | 6,353 | 6,109 | 96.2% | 5 | 285 | 335 | 69–914 |
| Both | 38 | 35 | 92.1% | 7 | 273 | 102 | 9–584 |
| Total | 12,753 | 6,935 | 54.4% | 14 | 236 | 360 | 77–976 |

Beyond these summary statistics, the full shape of each frame's posting-volume distribution is also worth inspecting directly, since a median alone can hide how spread out or skewed that distribution actually is. Fig. 2 shows the distribution of pre-index post volume for both discovery frames.

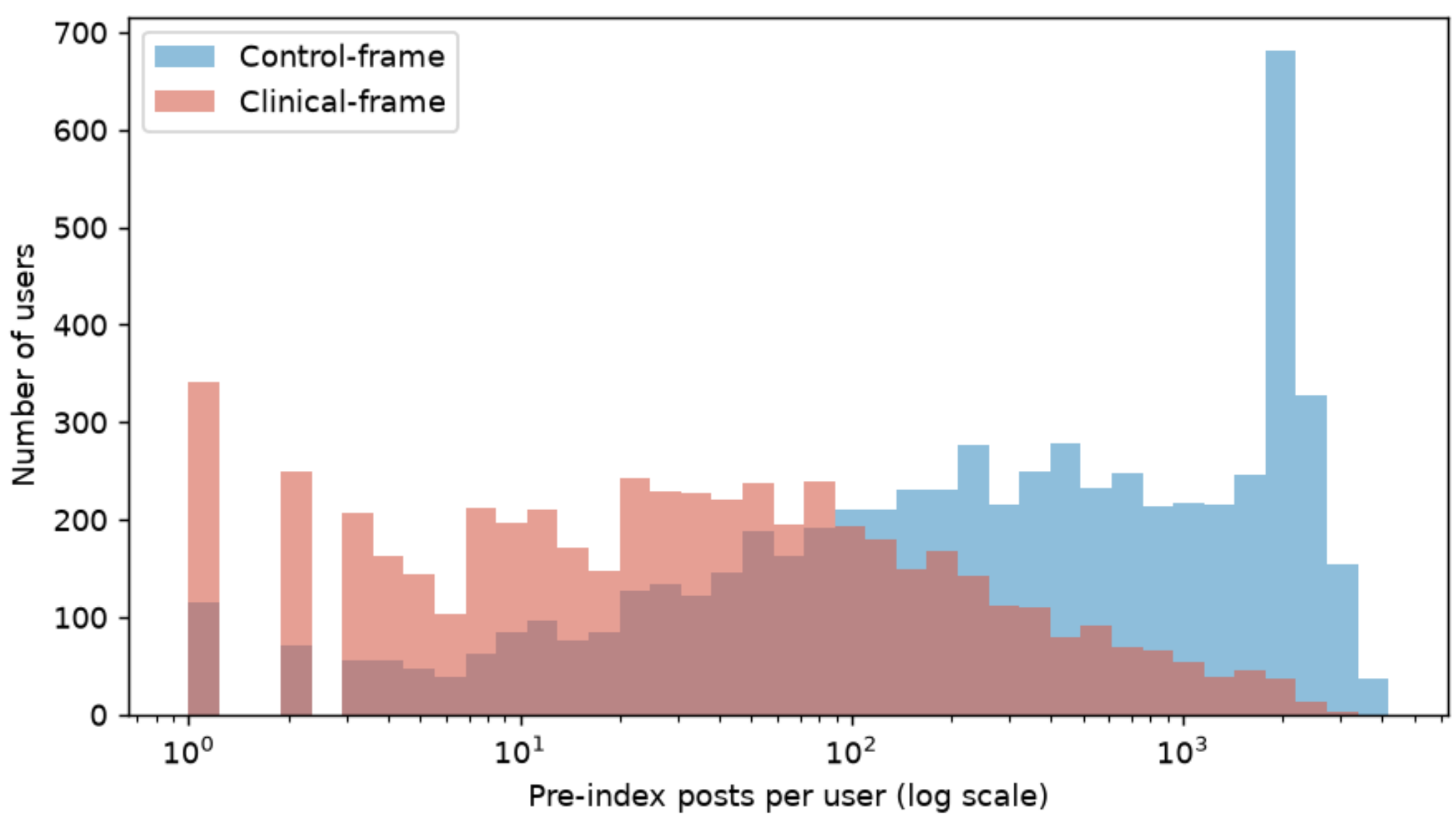

**Fig. 2.** Pre-index post volume per user for each discovery frame (log-scale histogram).

# 4. Methods

## 4.1. Feature extraction

All features are extracted from the pre-index post stream only. Text features concatenate `Title` and `Body` (using `Title` alone for link posts with null body).

### 4.1.1. TF-IDF text features

We fit a `TfidfVectorizer` [21] with English stop-word removal, unigrams (`ngram_range=(1,1)`), `min_df=5`, and `max_features=20,000` on training data only. We cap at 20,000 rather than a larger vocabulary because this gave stable convergence for the elastic-net Cox model within practical training time at our cohort size, with no further gain in preliminary checks against a larger vocabulary. All posts per user are concatenated into a single document prior to vectorisation (user-level representation).

### 4.1.2. MentalBERT embeddings

We encode each pre-index post independently using the `mental/mental-bert-base-uncased` checkpoint [20], a domain-adapted variant of BERT [22], extracting the `[CLS]` token embedding from the final layer with `max_length=128` and `fp16` mixed precision on a T4 graphics processing unit (GPU). This truncates only the small minority of posts longer than 128 tokens (mean post length is well under this threshold; Table 1); therefore, encoding is complete for typical posts but discards trailing content for unusually long ones. Per-user embeddings are obtained by mean-pooling the `[CLS]` vectors across all pre-index posts, producing a 768-dimensional user vector. Embeddings are computed once and cached to disk (checkpointed per batch to survive GPU session timeouts); all downstream training runs on a central processing unit (CPU).

### 4.1.3. Behavioural features

We compute six activity features from the pre-index post stream: (1) $\log(N_{\text{posts}} + 1)$ (posting volume); (2) posting rate per week (volume / observation span in weeks); (3) number of unique subreddits visited (community diversity); (4) mean upvote score across pre-index posts; (5) number of posts in the 90 days before the index time (recency burst); (6) number of days between the last pre-index post and the index time (recency gap). Feature (6) has a structural confound: since cases post right before transitioning, by definition, their recency gap is naturally shorter. We report a *robust* behavioural result that excludes this feature, in addition to the full-feature result; see Section 5.7.

## 4.2. Models

We evaluate four modelling approaches, chosen to isolate different sources of predictive signal: whether the model's functional form matters, whether non-linear interactions among behavioural features help, whether the order in which a user posts adds information, and whether the survival framing itself is doing any work relative to a standard classifier.

### 4.2.1. Cox proportional hazards regression

Cox proportional hazards regression [14] is a standard semi-parametric model for time-to-event outcomes. We use it as the primary model because it estimates the hazard of the event directly while naturally accommodating right-censored observations:

$$h(t \mid \mathbf{x}) = h_0(t)\exp(\boldsymbol{\beta}^\top \mathbf{x})$$

where $h_0(t)$ is the baseline hazard and $\mathbf{x}$ is the feature vector. For TF-IDF features, we first reduce dimensionality with Truncated SVD (singular value decomposition) ($k = 200$ components, latent semantic analysis (LSA)) before standardisation and Cox fitting. Behavioural features are standardised directly. Combined models concatenate the LSA-reduced TF-IDF vectors and standardised behavioural features into a single vector per user, so the Cox model weighs text-derived and behaviour-derived signals jointly rather than separately. MentalBERT embeddings (768-dim) are reduced with principal component analysis (PCA) ($k = 50$, retaining $\approx 85\%$ of training-fold variance) before Cox fitting. We verified that the TF-IDF Cox result is insensitive to this reduction: C-index is 0.763 with 200 components and 0.764 with 50, so the comparison between Cox and the Random Survival Forest baseline (both LSA-50) is not confounded by the difference in dimensionality.

Elastic-net regularisation [23,24] is applied (`CoxnetSurvivalAnalysis` from `scikit-survival`; $\alpha = 0.01$; $\ell_1$ ratio 0.5 for behavioural and combined models, 1.0 for TF-IDF). The regularisation strength (α) and $\ell_1$ ratio were fixed in advance at these standard values rather than tuned; keeping them fixed removes any risk of hyperparameter selection leaking across folds, though we make no claim that they are optimal.

#### 4.2.2. Random Survival Forest

Random Survival Forest (RSF) [18,25] is a non-parametric ensemble method for time-to-event prediction. We include it to test whether the linear form of the Cox model limits the behavioural signal, that is, whether non-linear interactions among activity features carry information a linear model cannot represent. Random Survival Forest models use $k = 50$ LSA components to reduce fitting time; the table labels these as LSA-50. Behavioural and combined features are standardised as for Cox. We fit RSF with 200 trees and a minimum of 5 samples per leaf (RandomSurvivalForest from scikit-survival), with other settings left at their library defaults.

#### 4.2.3. Sequential LSTM survival model

We include a Long Short-Term Memory (LSTM) network [26] to test whether the chronological order of a user's posts carries predictive signal beyond a static, mean-pooled text representation. For the sequential model (Section 5.8), we extract individual post embeddings rather than a single mean-pooled user vector. Each user is represented as a chronological sequence of up to 100 pre-index post embeddings (CLS tokens from MentalBERT, float16), zero-padded for shorter histories and processed oldest-first so the LSTM reads posts in temporal order. The model is a two-layer LSTM (hidden size 128, dropout 0.3) followed by a linear projection to a scalar log-hazard. Training minimises the Breslow-approximated Cox partial log-likelihood (the standard way of handling tied event times in the partial likelihood) using Adam (learning rate $10^{-3}$, weight decay $10^{-4}$) for 60 epochs with cosine annealing and gradient clipping at 1.0. Per-post embeddings are generated on a T4 GPU (Google Colab) and cached; all subsequent operations run on CPU.

#### 4.2.4. Logistic regression

Logistic regression is included as a non-survival baseline, to check that the survival framing is not costing us anything relative to plain classification. We fit it with the L-BFGS solver on TF-IDF and MentalBERT features; the behavioural result is reported in-text only, as a sanity check.

Table 2 below summarises the full experimental design: which combinations of feature type and model were tested, and why.

**Table 2.** Model × feature-type combinations tested in this study. ✓ (per-post) indicates MentalBERT embeddings used as an ordered per-post sequence rather than a single averaged vector; ✓ (in-text) indicates a result reported in the main text as a sanity check but not tabulated in Table 3/4; — = not tested.

| Model | TF-IDF | MentalBERT | Behavioural | Combined |
|---|---|---|---|---|
| Cox proportional hazards | ✓ | ✓ | ✓ | ✓ |
| Random Survival Forest | ✓ | — | ✓ | ✓ |
| LSTM (sequential) | — | ✓ (per-post) | — | — |
| Logistic regression (baseline) | ✓ | ✓ | ✓ (in-text) | — |

### 4.3. Evaluation metrics

- **Concordance index (C-index)** [16]: the primary discrimination metric; it measures the probability that, for a user having a shorter time to event than another user, the model correctly assigns higher risk. Random baseline is 0.5.
- **Precision-recall AUC (Area Under the Curve; PR-AUC)** [27]: secondary metric for the binary identification of eventual transitioners; appropriate for imbalanced data. Unlike C-index, PR-AUC's chance baseline is the event prevalence rather than 0.5, approximately 0.121 for Arm 1, so values should be read relative to that, not to 0.5. PR-AUC is not meaningful for Arm 2, since every user there is an event by construction. Accuracy is not reported. PR-AUC is computed against the binary event label, treating censored users as non-events. If censoring is informative, meaning censored users are not missing at random with respect to risk, this treatment causes the reported PR-AUC to underestimate the true PR-AUC.
- **Integrated Brier Score (IBS)**: measures calibration of survival probability estimates across the follow-up horizon; lower is better (null model ≈ 0.106 at our 12.1% event rate, computed as $p(1-p)$ where $p$ is the marginal event probability).

All interval metrics (C-index, PR-AUC, IBS) are reported as mean ± standard deviation across the five cross-validation folds, not as 95% confidence intervals.

### 4.4. Validation strategy

We use two complementary validation strategies:

#### 4.4.1. Grouped cross-validation

Five-fold `GroupKFold` [28] with `groups=User` ensures that no user appears in both training and test sets. This is the appropriate strategy because posts from a user share writing style, vocabulary, and risk status. A standard stratified or random split would inflate performance estimates by allowing train-test user leakage. Each test fold for Arm 1 contains approximately 1,267 users with a consistent event rate of approximately 12% (about 153 events per fold).

### 4.4.2. Temporal holdout

We train the models on users whose *first observed Reddit post* (discovery date) predates 2022-01-01 and test them on users whose discovery date falls on or after 2022-01-01 (i.e., 3,343 users in the training set; 2,994 users in the test set). This probes distribution shift across time (e.g., from platform demographic changes or language evolution) and is a stricter test than CV. The test period spans roughly 4.5 years (from January 2022 to the June 2026 cutoff); we refer to this throughout as the four-year holdout for brevity.

## 5. Results

### 5.1. Arm 1: Who will seek help? (RQ1)

We first address RQ1: which control-frame users eventually seek help. Table 3 reports discrimination (C-index, PR-AUC) and calibration (IBS) for every feature set under five-fold grouped cross-validation and the temporal holdout.

**Table 3.** Arm 1 (control-frame) results under five-fold grouped CV and the temporal holdout. Behavioural and combined models exclude the confounded recency-gap feature (full-feature variants in Table 7). IBS is reported for Cox text models only; RSF = Random Survival Forest; — = not computed.

| Model | CV C-index | CV PR-AUC | CV IBS | Holdout C-index |
|---|---|---|---|---|
| ***Text-based models*** | | | | |
| TF-IDF Cox | $0.763 \pm 0.018$ | $0.244 \pm 0.023$ | $0.093 \pm 0.004$ | 0.717 |
| TF-IDF LogReg | $0.715 \pm 0.008$ | $0.299 \pm 0.021$ | — | 0.711 |
| MentalBERT Cox | $0.758 \pm 0.018$ | $0.277 \pm 0.022$ | $0.089 \pm 0.005$ | 0.680 |
| MentalBERT LogReg | $0.699 \pm 0.026$ | $0.272 \pm 0.020$ | — | 0.661 |
| ***Behavioural and combined models (robust: recency-gap excluded)*** | | | | |
| Behavioural only | $0.797 \pm 0.018$ | $0.162 \pm 0.015$ | — | 0.846 |
| TF-IDF + Behavioural | $0.823 \pm 0.018$ | — | — | 0.856 |
| ***Sequential model*** | | | | |
| LSTM Cox (MentalBERT) | $0.790 \pm 0.012$ | 0.310 | — | — |
| ***Non-parametric baseline (Random Survival Forest)*** | | | | |
| RSF (TF-IDF, LSA-50) | $0.720 \pm 0.018$ | — | — | — |
| RSF (Behavioural, robust) | $0.941 \pm 0.005$ | — | — | — |
| RSF (Combined) | $0.895 \pm 0.010$ | — | — | — |

The TF-IDF text baseline ranks users well above chance, with a C-index of 0.763. Its calibration is only moderate, as the Integrated Brier Score of 0.093 sits close to the 0.106 expected from an uninformative model.

Domain adaptation does not help ranking. MentalBERT Cox matches TF-IDF on C-index (0.758) but edges ahead on precision-recall AUC (0.277 against 0.244) and on calibration (IBS 0.089 against 0.093). The two metrics reward different things. C-index measures how well a model orders users by risk, whereas PR-AUC measures precision in flagging the eventual transitioners. The domain-adapted embeddings therefore identify likely cases more precisely, while TF-IDF ranks the full population marginally better.

Behavioural features are the stronger signal. The robust behavioural model, which excludes the confounded recency-gap feature, reaches a C-index of 0.797, more than three points above TF-IDF (0.763), and its lead widens on the temporal holdout (0.846 against 0.717). A logistic regression on the same features gives a nearly identical precision-recall AUC (0.168) to the behavioural Cox model (0.162), so the survival framing costs nothing in precision relative to plain classification. Both figures are lower than the text-based models' PR-AUC (0.244–0.277), consistent with behavioural features being better at ranking overall risk than at pinpointing specific individuals (Section 6.2). Combining text and behaviour is best of all, reaching a C-index of 0.823 under cross-validation and 0.856 on the holdout. The full feature set that includes the confounded gap inflates these numbers and is reported only as a sensitivity check (Section 5.7). Fig. 6 visualises these C-index comparisons across both arms.

### 5.1.1. Non-parametric baseline (Random Survival Forest)

To test whether the linear form of Cox regression limits performance, we trained a Random Survival Forest [18,25] on the same feature sets. On text the forest does worse than Cox (C-index 0.720 against 0.763), so the linear bag-of-words structure suits the Cox partial likelihood better than a tree ensemble. On behaviour the picture reverses sharply. The forest reaches 0.941 against the behavioural Cox model at 0.797, and the combined forest reaches 0.895 against 0.823. This is an out-of-fold estimate (5-fold grouped cross-validation, SD 0.005 across folds); the corresponding in-sample C-index is 0.985, a gap consistent with the ensemble's higher flexibility relative to the linear Cox baseline rather than fold leakage, given the tight out-of-fold variance across folds. This gap points to strong non-linear interactions among the behavioural features that the proportional hazards form of Cox cannot capture (Section 7.3). Permutation importance (Fig. 3) traces most of this signal to sheer activity. Posting volume and posting rate dominate, with C-index drops of 0.35 and 0.19 when permuted, while recency burst, upvote score, and community diversity each contribute little, with drops of 0.05 or less.

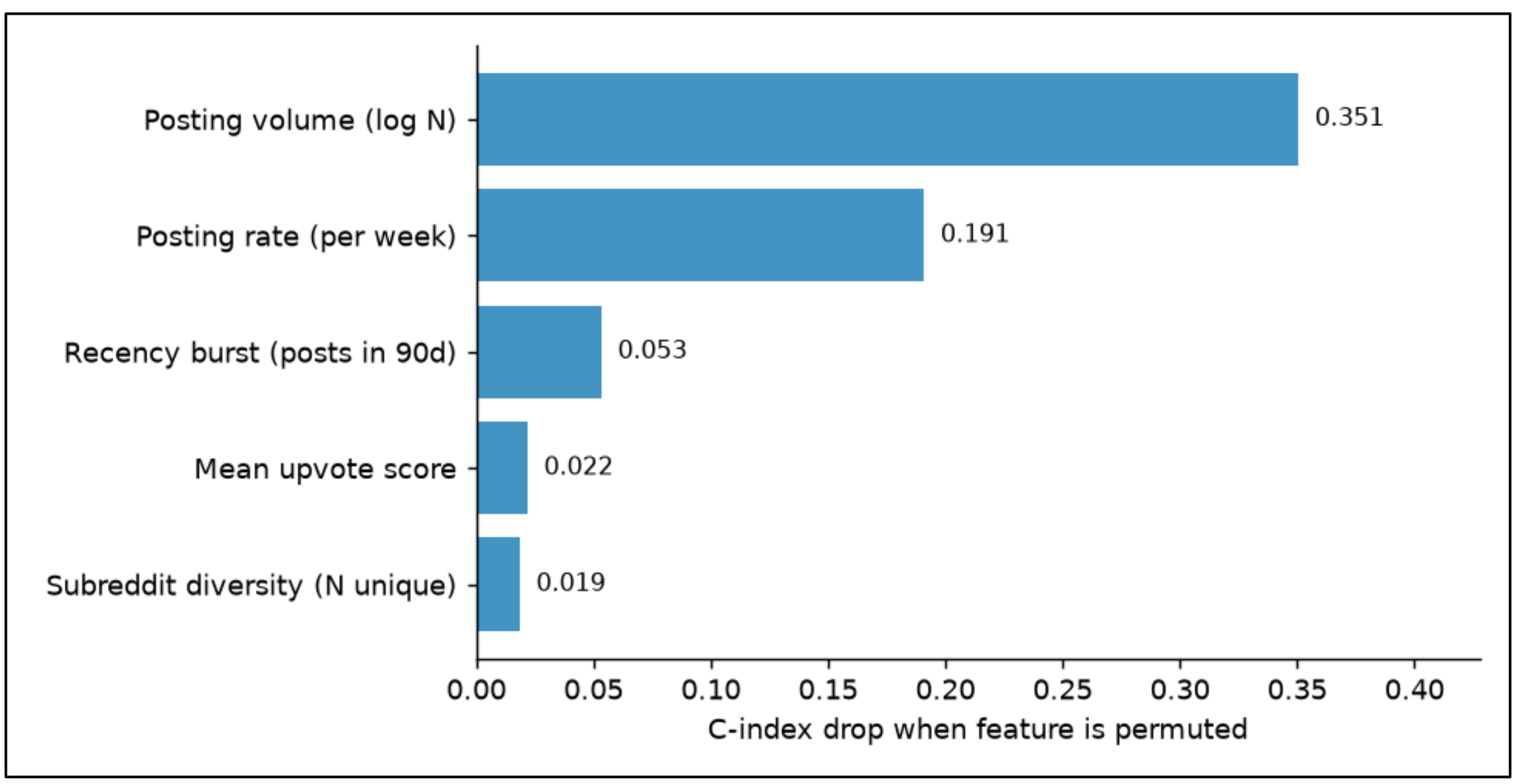

**Fig. 3.** Permutation feature importance for the behavioural Random Survival Forest (Arm 1): C-index drop when each feature is permuted.

Fig. 4 shows the correlation structure among these behavioural features. Total posts, posting rate, and their log-transform are highly collinear (ρ up to 1.00), while days since the last pre-index post is the feature most strongly associated with the event indicator (ρ=−0.27), consistent with the recency-based signal explored in Section 5.5.2.

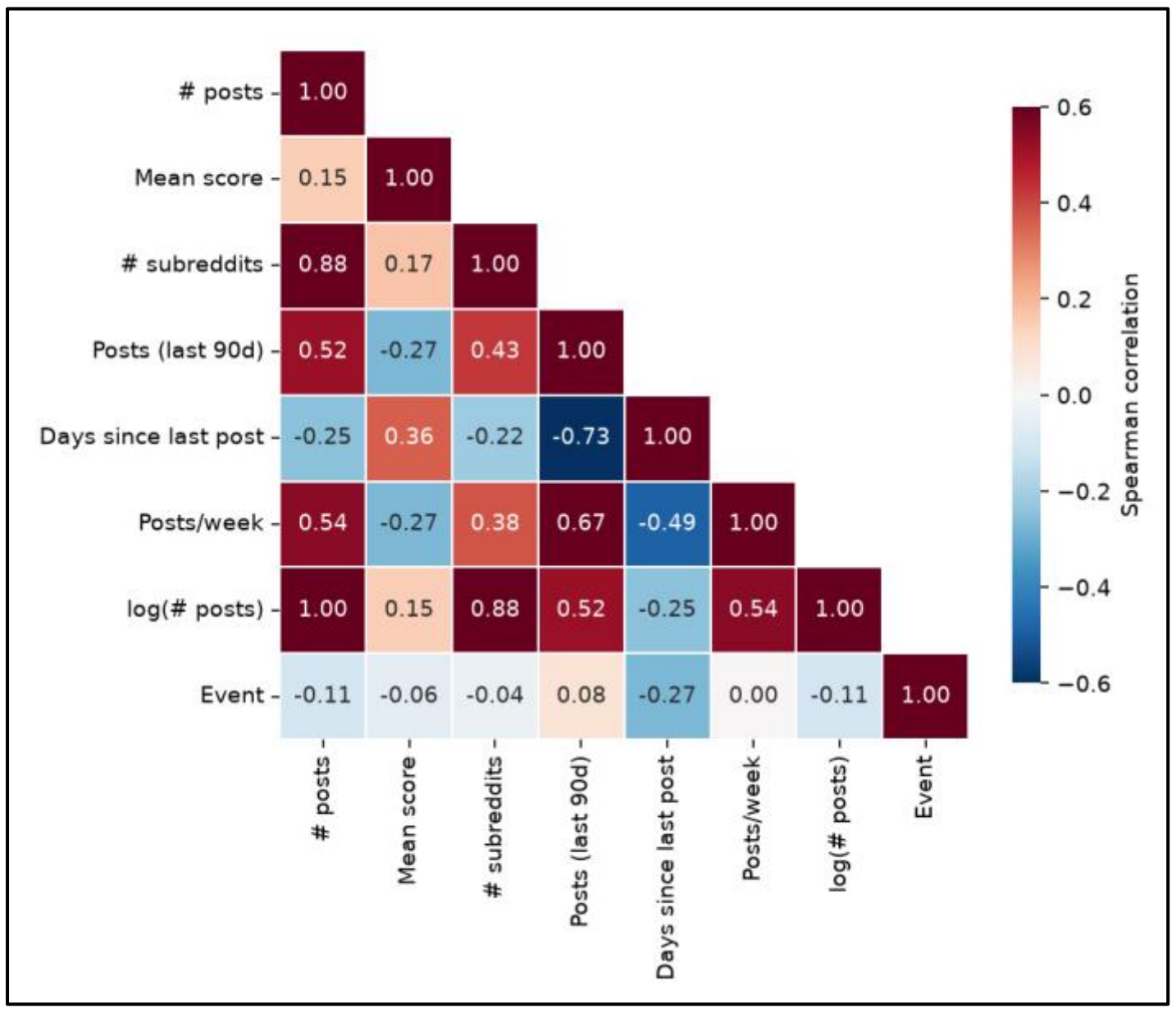

**Fig. 4.** Spearman correlation among behavioural features and the event indicator (Arm 1).

As a robustness check on these behavioural features, we also examined whether transitioners and censored users differ in when they post, independent of how much or how diversely they post. Although this was not a focus of the main analysis, it rules out a simple confound: if the two groups posted at systematically different times of day, the behavioural signal above could partly reflect circadian differences rather than genuine engagement or volume effects. Fig. 5 shows no such difference: posting-time distributions are close to indistinguishable between the two groups; therefore behavioural predictors are not standing in for a time-of-day artefact.

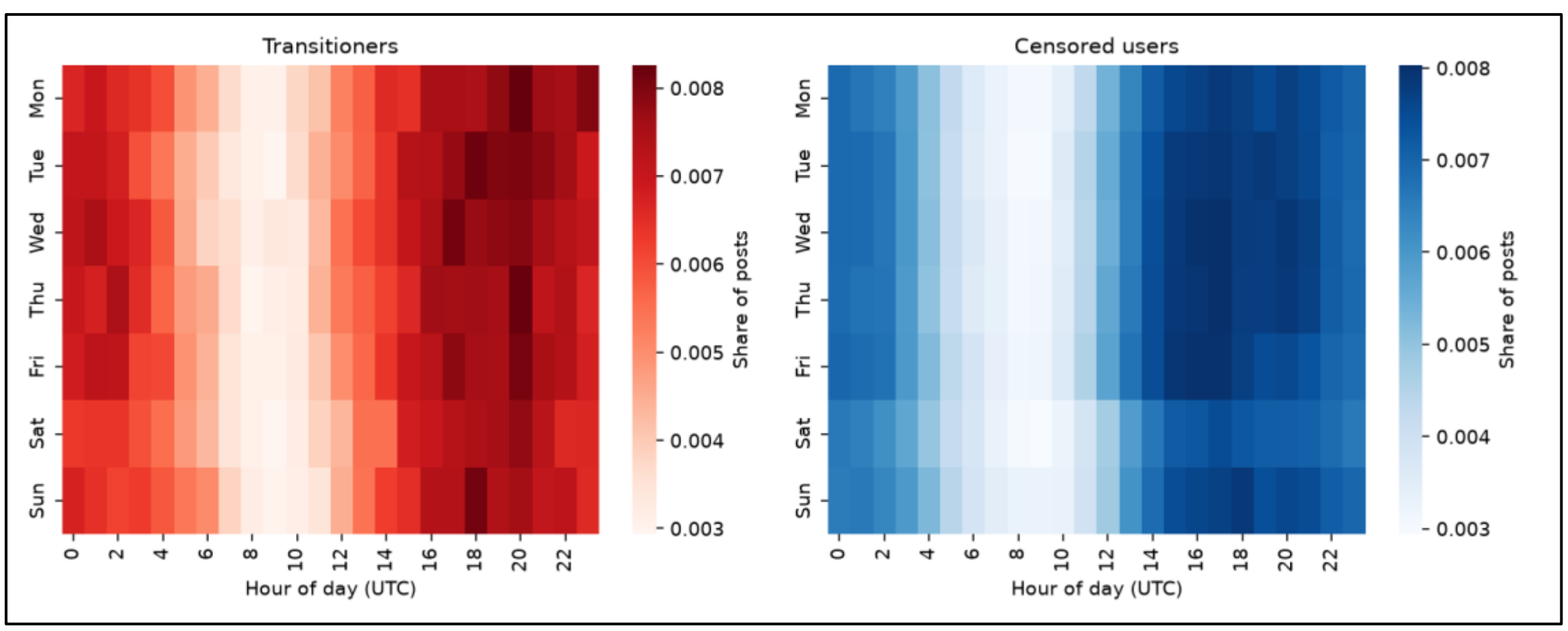

Fig. 5. Share of pre-index posts by hour and day of week (UTC), transitioners vs. censored users (Arm 1); included to check that behavioural predictors are not confounded by posting-schedule differences.

### 5.1.2. Temporal holdout

The temporal holdout tells a sharper story. Both text models lose ground on the post-2022 users, TF-IDF Cox falling to 0.717 (a drop of about 6% from cross-validation) and MentalBERT Cox to 0.680 (about 10% lower). The behavioural model moves the other way, improving to 0.846, and the combined model rises to 0.856 from 0.823. Activity-based signals generalise across years better than language, which suggests they capture more time-invariant aspects of help-seeking behaviour.

## 5.2. Arm 2: How imminent is the transition? (RQ2)

We now turn to RQ2: among clinical-frame users, who all eventually seek help, can pre-clinical language rank how imminent each transition is? Table 4 reports imminence C-index for each feature set.

**Table 4.** Arm 2 (clinical-frame; n=5,023; all events; median lead 335 days): imminence prediction. All users eventually transitioned, so models rank them by how soon. Behavioural C-index is unstable across folds (see text).

| Model | C-index | Stability |
|---|---|---|
| TF-IDF only | $0.698 \pm 0.003$ | Stable |
| Behavioural only | $0.384 \pm 0.231$ | Unstable (not reported) |
| TF-IDF + Behavioural | $0.589 \pm 0.116$ | Moderate |
| LSTM Cox (MentalBERT) | $0.667 \pm 0.003$ | Stable |
| ***Non-parametric baseline (RSF)*** | | |
| RSF (TF-IDF, LSA-50) | $0.683 \pm 0.005$ | Stable |

On Arm 2, TF-IDF text ranks users by imminence at a stable C-index of 0.698, well above the chance level of 0.5 and consistent across folds. Because every user in this arm eventually transitioned, the number has a concrete reading. For any two users who will seek help, the model picks the more imminent one about 70% of the time, using nothing but the content of their pre-clinical posts.

Behavioural features fail here. Their C-index of 0.384 sits near chance and swings widely across folds, a sign of structural instability. The features that reveal who will transition in Arm 1 say almost nothing about how soon in Arm 2. That split is itself a result. The likelihood of transitioning is best read from how people behave, whereas timing is written into what they say; therefore the two arms capture complementary parts of the help-seeking process. The text forest on Arm 2 (0.683) matches text Cox (0.698), which again shows that a simple linear bag-of-words model already extracts essentially all the timing signal the text contains.

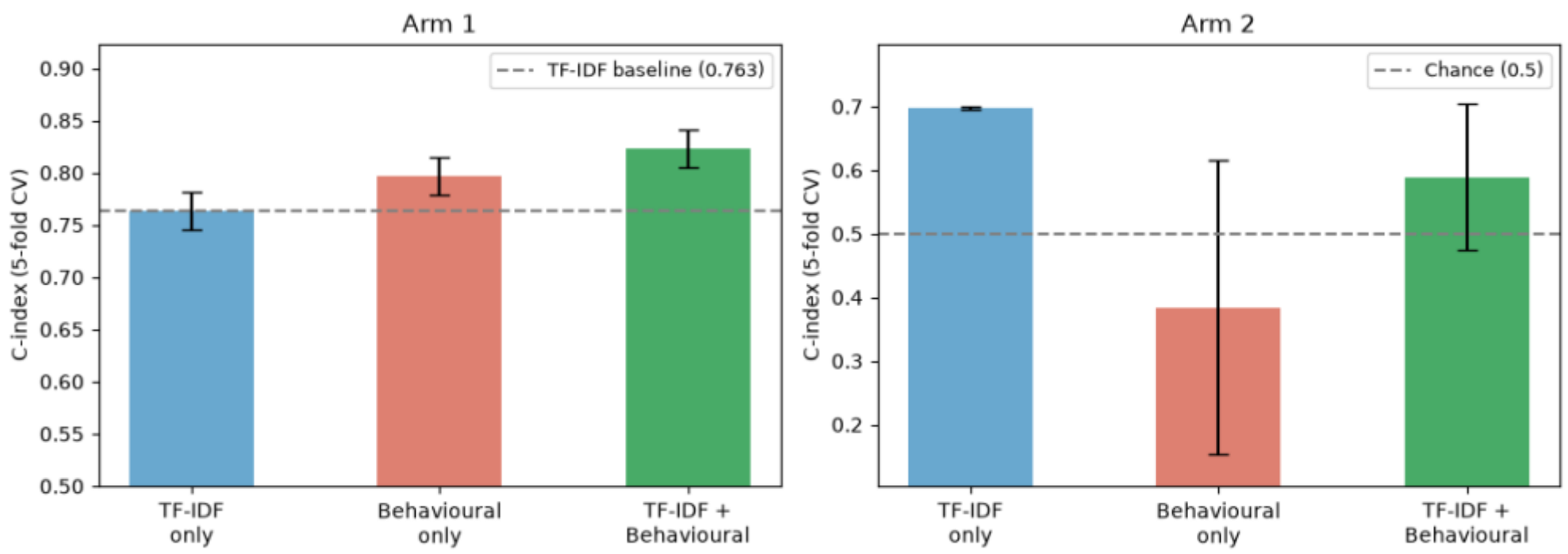

**Fig. 6.** C-index across feature sets for both arms (5-fold CV; error bars ±1 SD, standard deviation).

## 5.3. Kaplan–Meier lead-time curve

To characterise when transitions occur in the control frame, we estimate the time-to-event distribution non-parametrically with the Kaplan–Meier method. The Kaplan–Meier curve in Fig. 7 falls steadily, with transitions spread across the whole observation window rather than clustered at any point. Most users never transition, with 87.9% remaining censored, which matches how uncommon formal online help-seeking is at the population level.

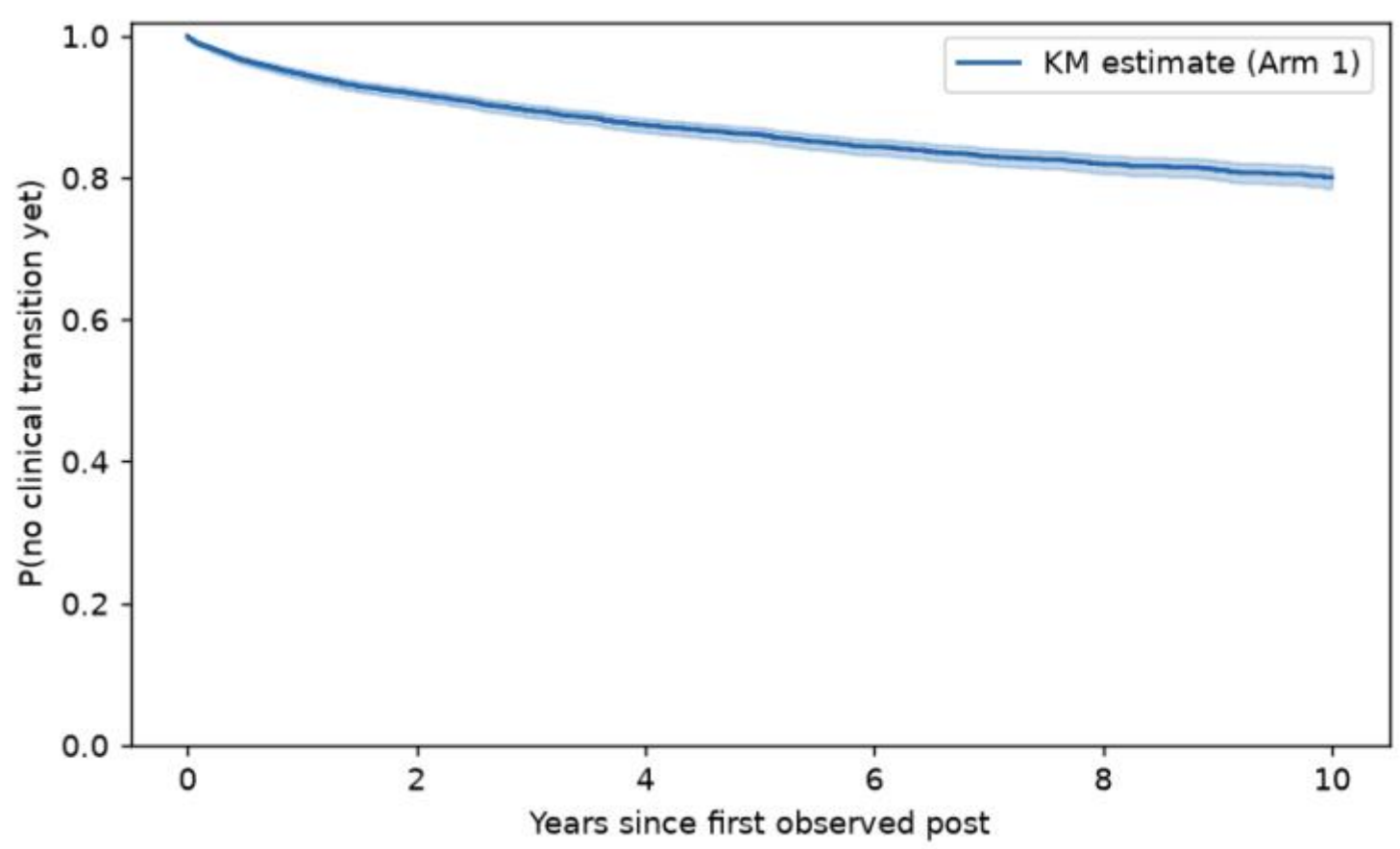

**Fig. 7.** Kaplan–Meier estimate of time to first clinical post (Arm 1; shaded band = 95% CI, confidence interval).

To gauge clinical usefulness, Fig. 8 stratifies the same curve by predicted risk quartile from the behavioural forest, with scores computed out of fold to avoid in-sample optimism. Observed transition rates climb from 1% in the lowest-risk quartile to roughly 20% in the top two, so the model separates a large low-risk majority before anyone transitions. The top two quartiles resemble each other because out-of-fold risk scores are noisy, but that does not weaken the clear gap between the lowest quartile and the rest.

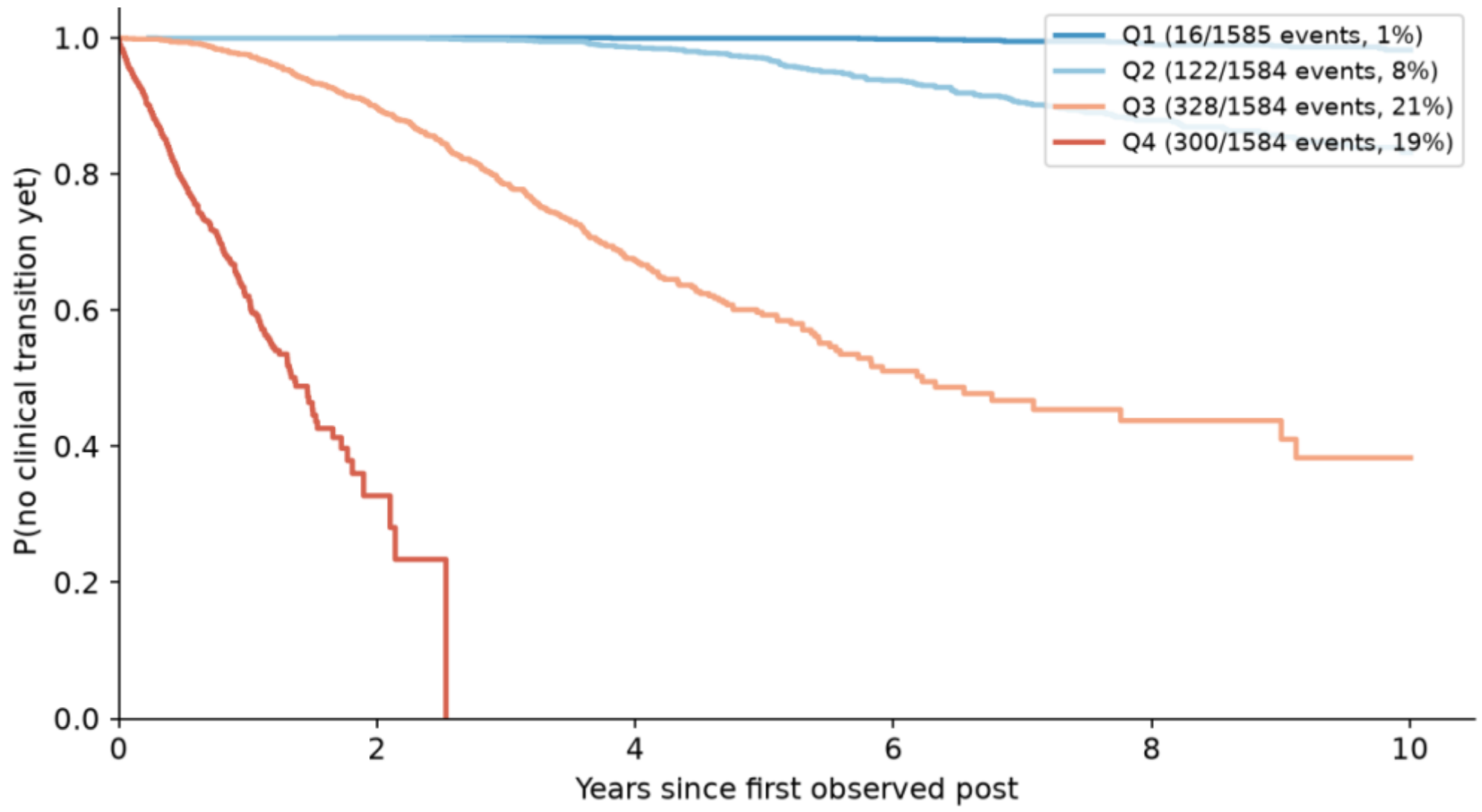

**Fig. 8.** Kaplan–Meier curves by predicted risk quartile (behavioural RSF, Arm 1; risk scores from 5-fold CV).

## 5.4. Linguistic signals

Comparing mean TF-IDF weights between eventual transitioners and censored controls (Table 5) reveals a clear thematic split. The terms that mark future cases are emotional and introspective, such as feel, anxiety, and life, while those that mark controls are technical and casual, dominated by URL and image-format tokens such as jpeg, png, and gif, along with filler like yeah. The pattern mirrors the familiar contrast between expressive, narrative posting tied to emotional distress and informational or media sharing tied to casual browsing [4,6].

**Table 5.** Top terms by mean TF-IDF weight difference (cases minus controls). Positive: enriched in users who eventually transitioned. Negative: enriched in censored controls.

| Enriched in cases (+) | | Enriched in controls (−) | |
|---|---|---|---|
| **Term** | **Δ** | **Term** | **Δ** |
| know | +0.014 | https | −0.022 |
| feel | +0.013 | redd | −0.012 |
| really | +0.010 | webp | −0.011 |
| want | +0.008 | jpeg | −0.009 |
| help | +0.008 | gif | −0.008 |
| anxiety | +0.006 | com | −0.008 |
| life | +0.006 | yeah | −0.007 |
| friends | +0.006 | png | −0.007 |

Fig. 9 visualises this contrast directly, after additionally removing the URL and image-hosting tokens already documented in Table 5 (e.g. jpeg, png, webp), to isolate the genuine conversational vocabulary underneath. Even with that link-sharing artefact removed, the split persists: transitioners' language is dominated by emotional and help-seeking vocabulary, while censored users' language is dominated by casual, entertainment-oriented content. Therefore, the difference between both groups is not merely that controls share more links.

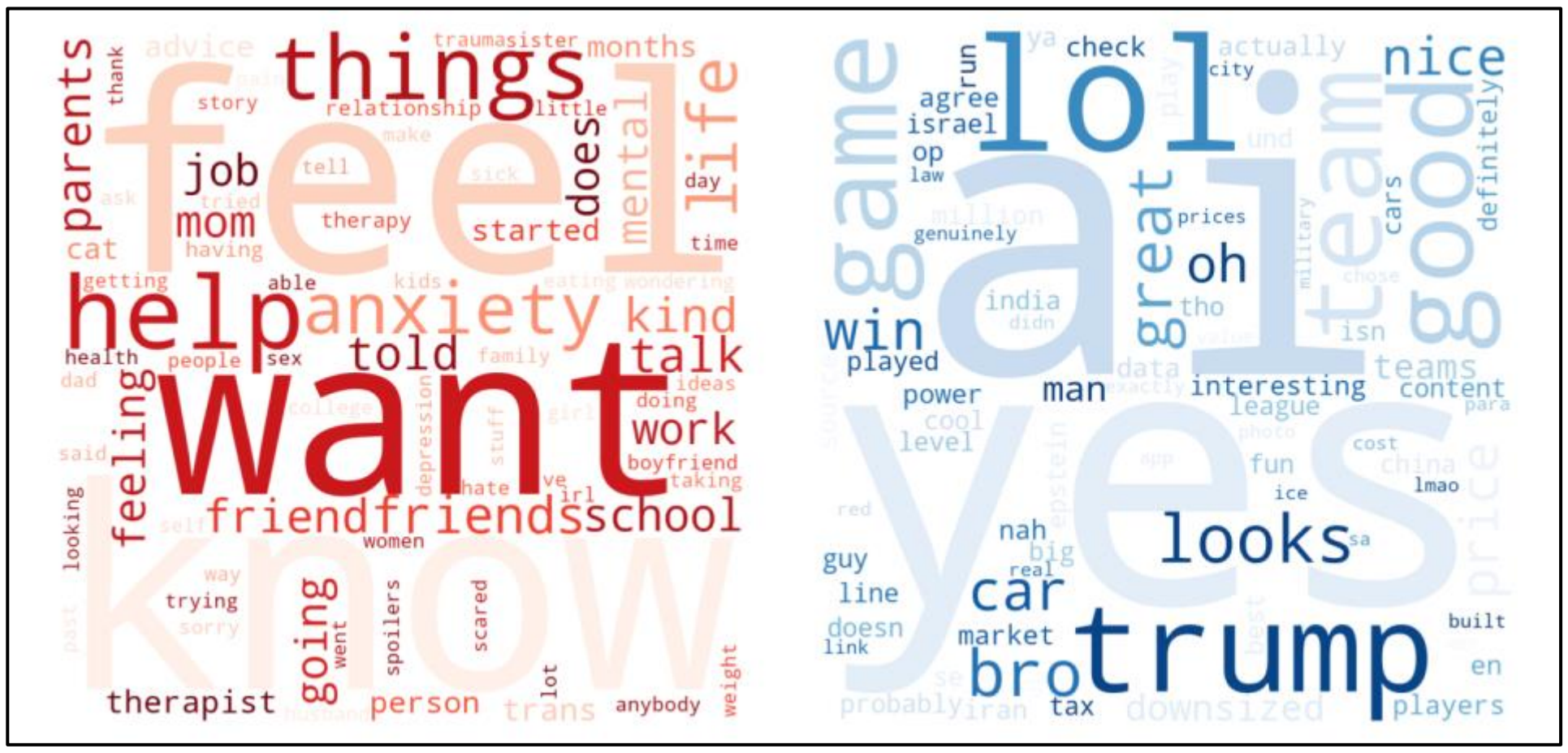

**Fig. 9.** Word clouds of TF-IDF terms most enriched among transitioners (left) and censored users (right) after removing the URL/image-hosting tokens shown in Table 5, so that both reveal the underlying conversational vocabulary; term size reflects the mean weight difference.

## 5.5. Temporal trajectory: early vs late window

We investigate whether language content *changes* as a user approaches a clinical transition, testing the hypothesis that vocabulary shifts toward more emotionally expressive or help-seeking content near the event.

### 5.5.1. Experiment A: temporal window split

We divide each user's pre-index history at the temporal midpoint (median post timestamp), producing an "early" half and a "late" half. Table 6 compares TF-IDF Cox models trained on each half and their combination.

**Table 6.** Early versus late temporal window (Arm 1). The full-history model uses all pre-index posts.

| Feature set | C-index (5-fold CV) |
|---|---|
| All posts (full history) | $0.763 \pm 0.018$ |
| Early half only | $0.742 \pm 0.017$ |
| Late half only | $0.770 \pm 0.016$ |
| Early + late combined | $0.772 \pm 0.015$ |

Language closer to the transition carries more signal. The recent half of each user's history ranks better than the early half (C-index 0.770 against 0.742; Fig. 10), consistent with a gradual shift toward more emotionally expressive or help-seeking content as the transition nears. Using both halves together does slightly better still (0.772 against the full-history baseline of 0.763), which suggests the early and late periods hold partly distinct signal rather than repeating the same vocabulary. This is not simply a difference in data: 'full history' pools all pre-index posts into a single vectorised document, which discards when within the window a term appeared, whereas 'early+late combined' fits two separate vectors, one per half, and concatenates them, preserving that temporal placement. It is this structural difference, not additional information, that the combined representation exploits.

### 5.5.2. Experiment B: recency-weighted TF-IDF

We also weighted each post's TF-IDF vector by $w_t = \exp(-\lambda \cdot d_t/365)$ where $d_t$ is the number of days before the index time, and $\lambda \in \{0, 0.25, 0.5, 1.0, 2.0\}$. Fig. 10 shows the C-index as a function of $\lambda$.

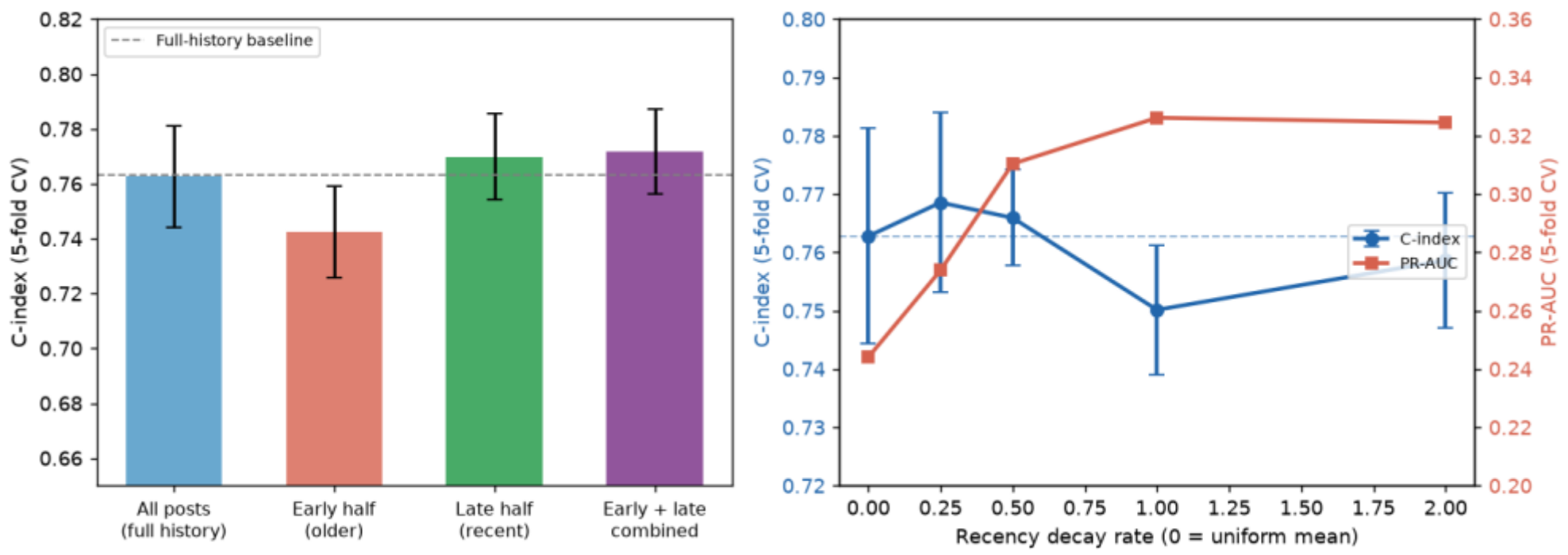

**Fig**. 10. **Temporal-window split (left) and recency-weighting sweep (right, C-index and PR-AUC) for TF-IDF (Arm 1).**

Recency weighting helps precision more than ranking. Down-weighting older posts lifts the C-index only slightly, from 0.763 to the best value of 0.769 at a decay rate of 0.25, confirming that language nearer the index time carries somewhat stronger signal. Precision-recall AUC, by contrast, climbs much more, from 0.244 at no weighting to 0.310 at a rate of 0.5 and 0.326 at 1.0 before it plateaus. Aggressive recency weighting therefore sharpens the model's ability to flag specific high-risk individuals more than it improves overall ranking. For screening uses that value precision over ranking, a stronger weighting near 1.0 may be preferable to uniform pooling.

Together, these experiments establish that the help-seeking signal is *distributed across the full pre-index history* but that recent posts carry disproportionate precision signal. The sequential LSTM model (Section 5.8) tests whether the *order* of that history carries information beyond recency-weighted aggregation.

## 5.6. Most-recent-*K* posts ablation

We test whether restricting features to the $K$ most recent pre-index posts per user (rather than the full history) improves performance. This ablation addresses the concern that early posts, which may predate any mental health difficulties by years, could dilute the signal in the user's representation.

Performance rises steadily as more of each user's history is included, from a C-index of 0.677 when only the 10 most recent posts are used to 0.763 when the full history is used (Fig. 11). This clearly shows that older posts add signal rather than noise. The help-seeking signal is therefore spread across a user's entire posting record rather than concentrated in recent activity, echoing the strength of total posting volume among the behavioural features. In practice, restricting prediction to recent posts would throw away useful information. Therefore models should use the full pre-index history whenever it is available, up to the limit imposed by the Reddit API.

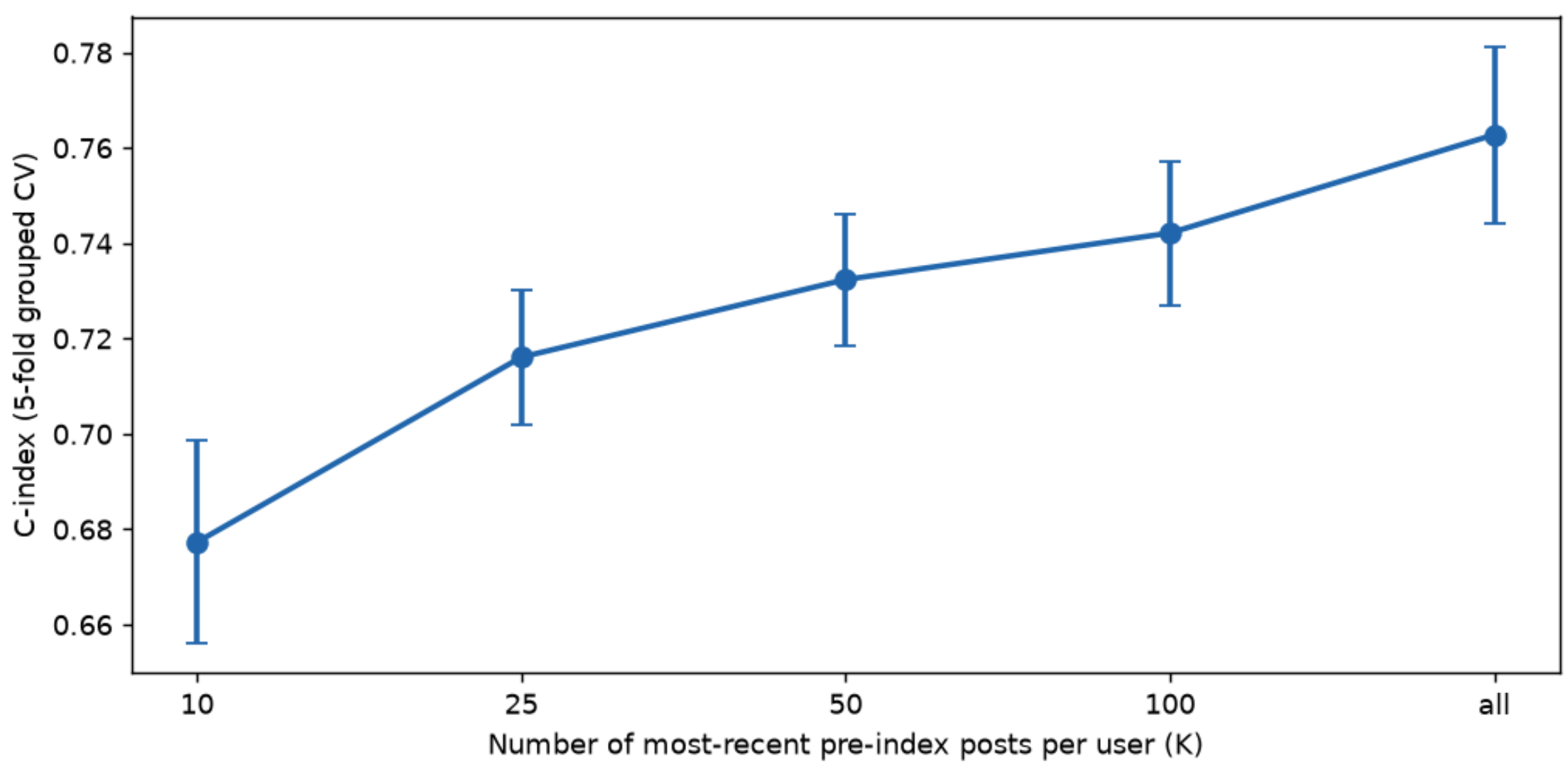

**Fig. 11.** C-index versus the number of most-recent posts used per user (Arm 1).

## 5.7. Sensitivity analysis

The recency-gap feature, i.e. the number of days between a user's last pre-index post and the index time, carries a structural confound. By definition, as cases post right up to their transition, their gap is short; whereas censored users may have gone quiet months or years before the cutoff. The worry is that the model could be reading this timing artefact rather than genuine behaviour. Table 7 shows that removing the feature reduces the behavioural advantage but does not erase it.

**Table 7.** Sensitivity analysis: behavioural Cox C-index under progressive feature ablation. CV: five-fold grouped cross-validation. Holdout: train pre-2022, test 2022+. Feature 5 = posts_last_90d (recency burst); Feature 6 = days_since_last (recency gap, structurally confounded).

| Feature set | CV C-index | Holdout C-index |
|---|---|---|
| Full behavioural (incl. burst + gap) | $0.823 \pm 0.014$ | — |
| Robust (excl. gap only) | $0.797 \pm 0.018$ | 0.846 |
| Ablation (excl. burst + gap) | $0.795 \pm 0.019$ | — |
| TF-IDF (reference) | $0.763 \pm 0.018$ | 0.717 |

The behavioural advantage survives aggressive ablation. Removing both the confounded recency gap and the recency burst leaves a behavioural C-index of 0.795, still comfortably above the text baseline of 0.763, so the result is not an artefact of recency. The tiny difference between the robust model (0.797) and this further ablation (0.795) shows that the burst feature adds little beyond volume and diversity. Keeping the confounded gap inflates the behavioural model's cross-validation score to 0.823, with a similar inflation for the combined model, which is why we treat both as unreliable and report the robust figures of 0.797 and 0.823 throughout. What remains is carried mainly by posting volume, subreddit diversity, and mean engagement, matching the permutation ranking in Fig. 3.

## 5.8. Sequential LSTM model

We trained a sequential LSTM survival model over per-post MentalBERT embeddings (Section 4.2.3). Results are reported in Table 3 (Arm 1) and Table 4 (Arm 2).

Post order carries information. On Arm 1, the sequential LSTM reaches a C-index of 0.790, ahead of both static TF-IDF Cox (0.763) and mean-pooled MentalBERT Cox (0.758). Modelling how a user's language evolves across their history, rather than collapsing it into a single averaged vector, gives a measurable gain in ranking. This is the clearest evidence that the trajectory of posts, not just their aggregate content, helps predict who will seek help.

On Arm 2, the ordering does not help. The LSTM reaches only 0.667 ± 0.003 (5-fold grouped CV), below the TF-IDF Cox baseline of 0.698, echoing the recency-weighting result. For imminence, what users say matters more than the order in which they said it, and a simple bag-of-words model already ranks users by proximity to transition about as well as the sequence model does.

Together with the recency-weighting experiments, these results establish a consistent picture: sequential structure in posting history matters for predicting *who* will eventually seek help (Arm 1), but not for predicting *how soon* (Arm 2), where the lexical content of the pre-index period is the dominant signal regardless of its temporal arrangement.

## 5.9. Calibration analysis

Good ranking does not guarantee good probabilities. Fig. 12 (MentalBERT LogReg, temporal holdout) and Fig. 13 (combined Cox, cross-validated) both sit below the perfect-calibration diagonal, so both models systematically overestimate the chance of transition. This is expected when the event rate is low (here 12.1%) and no explicit calibration step is applied. It also shows that discrimination and calibration are separate properties, since a model can order users well while its absolute probabilities are off. The right panel of Fig. 13 makes the ranking point directly. The combined model assigns clearly higher risk to eventual transitioners than to censored users, in line with its C-index of 0.823, even though those probabilities would need recalibration before any real-world use.

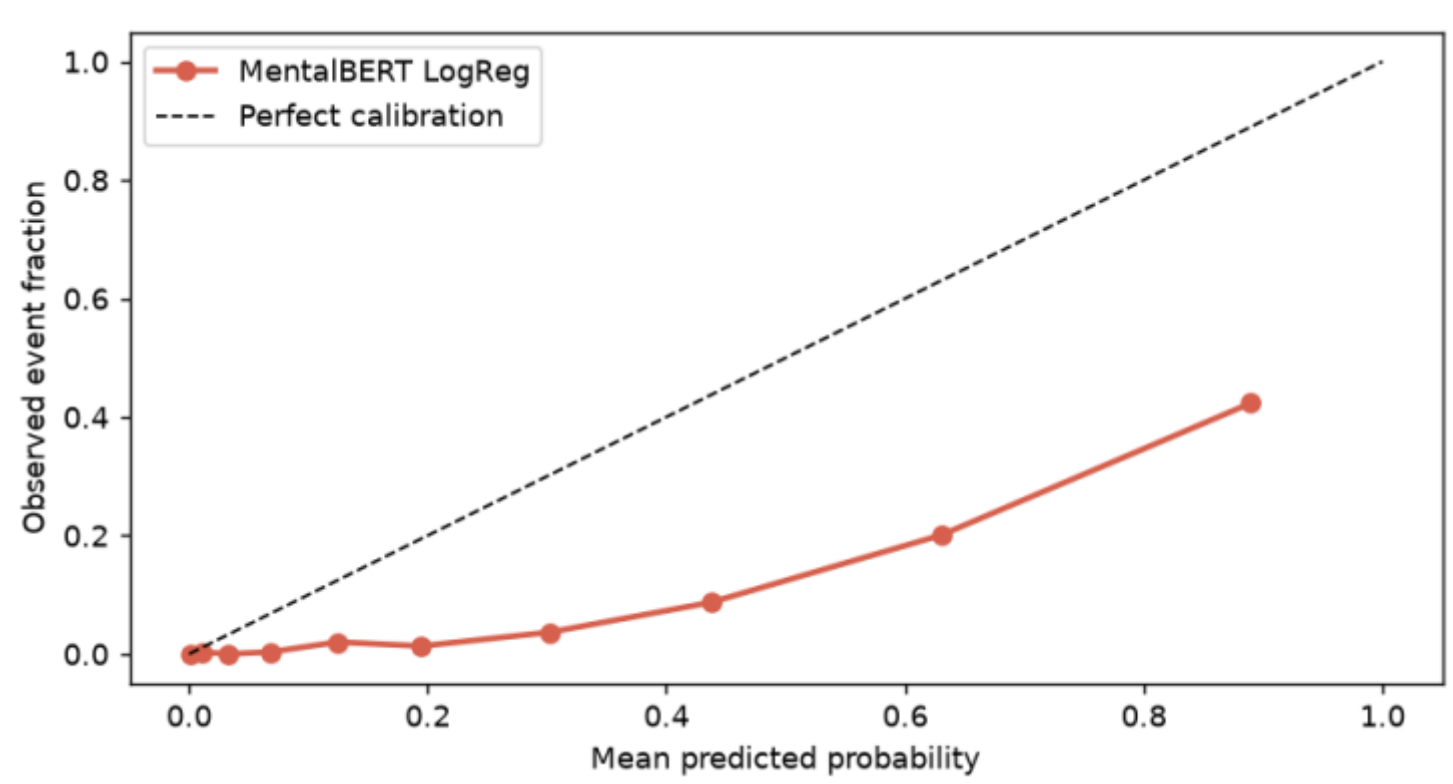

**Fig. 12.** Calibration of the MentalBERT baseline on the temporal holdout (Arm 1).

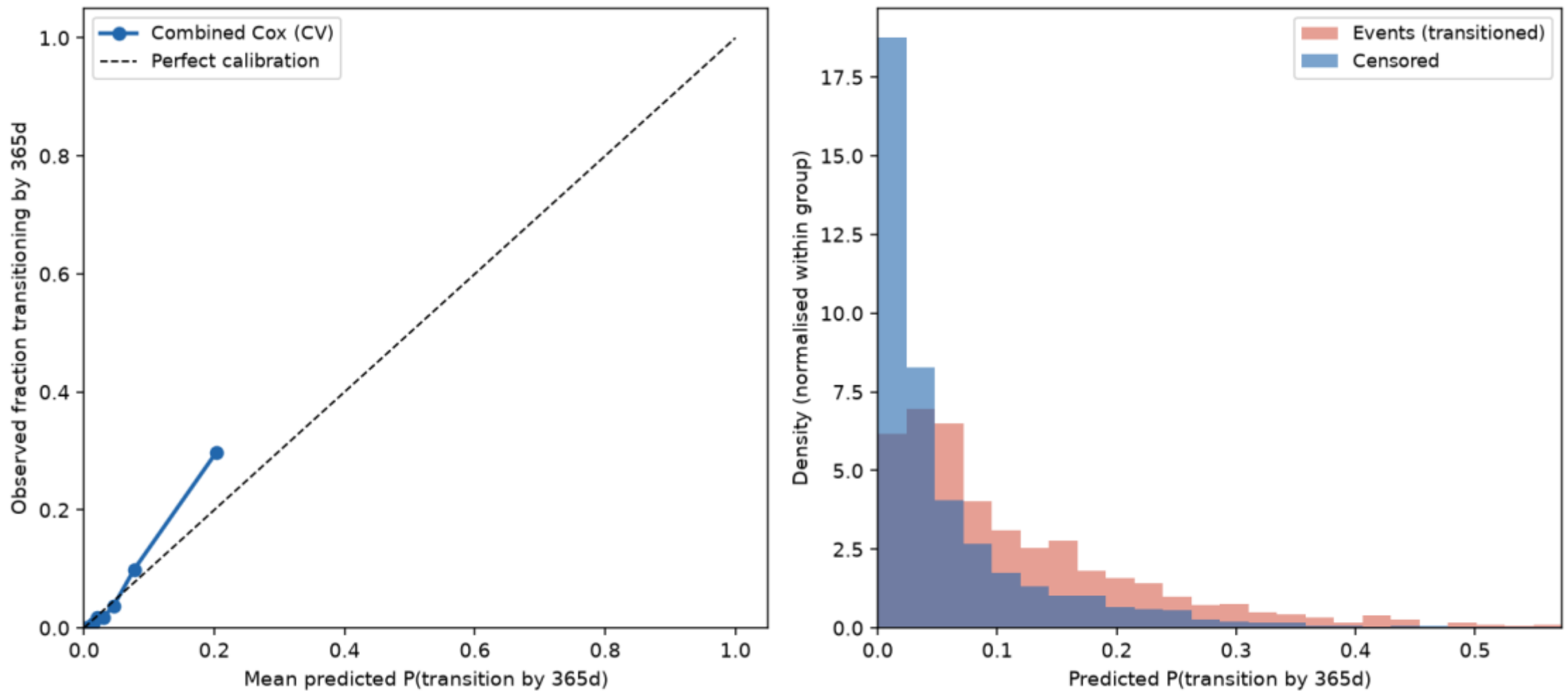

**Fig. 13.** Combined-model calibration (left) and risk-score distribution by outcome (right, densities normalised within each group so the comparison is not swamped by class imbalance), Arm 1.

Fig. 14 makes the class imbalance concretely visible: at the conventional 0.5 threshold the combined model predicts almost no transitions within a fixed 365-day horizon, while thresholding at the cohort base rate (6.0% — the one-year cumulative incidence among the N=4,902 users with an observable 365-day outcome, lower than the full-cohort 12.1% eventual transition rate since it excludes both the longer tail of later transitions and users censored before one year) recovers most transitioners at the cost of many false positives, underscoring why threshold-free ranking metrics such as the C-index are reported as the primary measure throughout (Section 4.3).

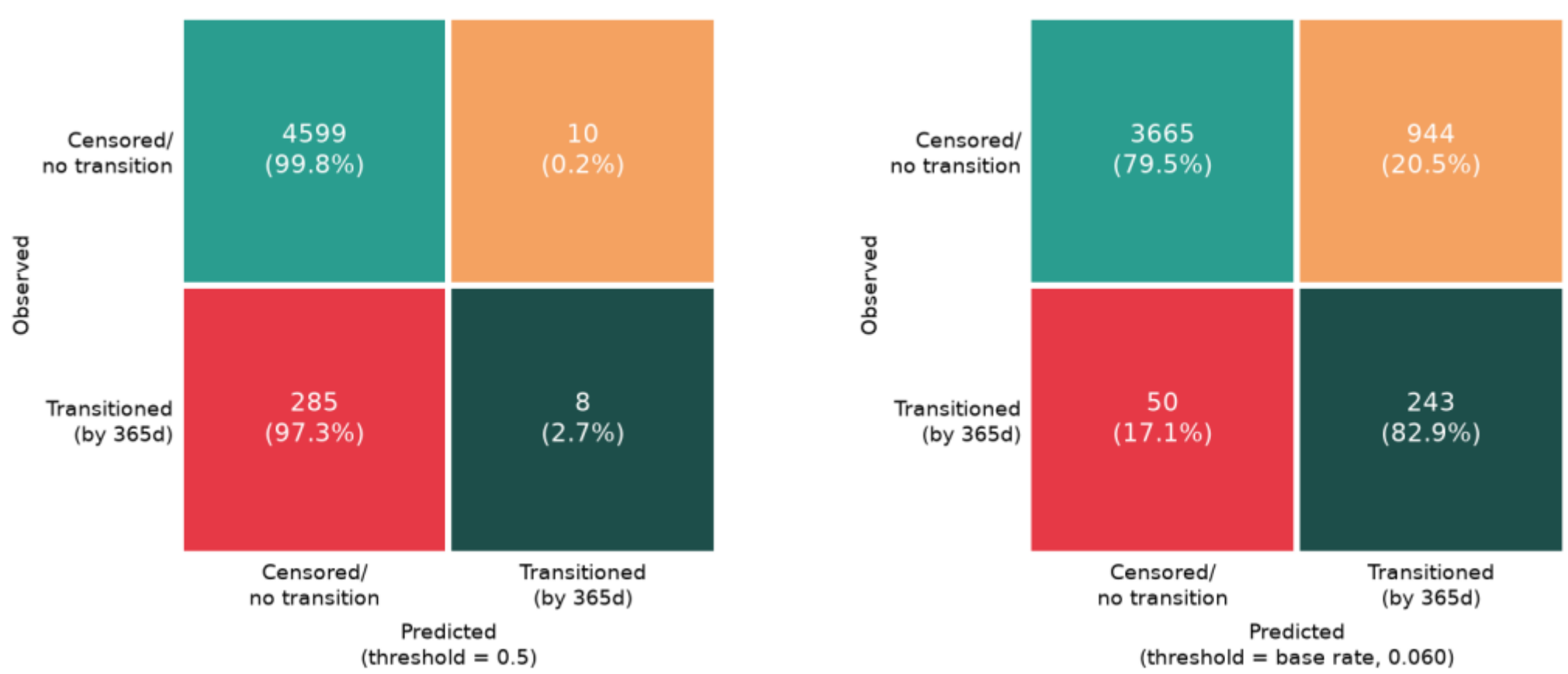

**Fig. 14.** Confusion matrices for the combined Cox model at a fixed 365-day horizon (5-fold cross-validated, Arm 1), at a threshold of 0.5 (left) and at the cohort base rate (right).

## 6. Discussion

### 6.1. Answering the research questions

**RQ1 (who?):** Pre-transition features can identify which users will eventually post to a clinical community, with a C-index of 0.823 (with the combined model) and 0.797 (with behavioural features alone), both robust to the confounded recency-gap feature. Behavioural features (e.g., posting volume, community diversity) are the dominant predictors. The result strengthens under a four-year temporal holdout (with a C-index of 0.856 for the combined model and 0.846 for the behavioural features), confirming generalisability across time.

**RQ2 (how soon?):** For users who are known to transition, text features (i.e., TF-IDF Cox) provide stable imminence prediction (with a C-index of 0.698). Behavioural features do not transfer to imminence prediction (C-index of 0.384, unstable). Post order (i.e., using LSTM) adds no advantage over aggregated text for this task (C-index of 0.667).

### 6.2. Why do behavioural features outperform text?

Two complementary explanations are plausible. First, posting volume and community diversity are proxies for *platform engagement*: users who participate broadly on Reddit are more likely to eventually encounter and join a clinical community. Second, increased activity and diversity may reflect *life transitions and social seeking*: individuals approaching mental health help-seeking may show elevated online activity as they process experiences and seek information across communities.

Both explanations are not mutually exclusive, and the PR-AUC result is informative: behavioural features rank users more accurately (with higher C-index) but identify specific individuals less precisely (with lower PR-AUC: 0.162 for the full behavioural Cox vs. 0.244 for TF-IDF Cox). This suggests activity patterns capture *baseline risk* better than text, while specific language content is more diagnostic of *individual propensity*. The Random Survival Forest achieves C-index 0.941 on behavioural features alone, substantially exceeding the Cox model (0.797), which indicates strong non-linear interactions among activity features that a linear proportional hazards model cannot represent (which is confirmed by the Schoenfeld test of the proportional-hazards assumption; see Section 7.3).

This engagement account is visible directly from the data. Fig. 15 plots each subreddit's transition rate among its visitors against its visitor count: niche advice, identity, and support-oriented communities (e.g. r/findapath, r/CPTSD, r/DecidingToBeBetter) sit well above the cohort baseline despite modest traffic, while large general-interest subreddits cluster close to the cohort baseline. The behavioural Random Survival Forest itself uses only an aggregate diversity count, not which communities a user visits, so this pattern is not a feature the model sees directly; it is evidence for why that aggregate count carries signal in the first place.

### 6.3. Why does MentalBERT not improve C-index?

TF-IDF and MentalBERT are tied on C-index (0.763 vs. 0.758) but MentalBERT numerically outperforms TF-IDF on PR-AUC (0.277±0.022 vs. 0.244±0.023) against a chance baseline of ≈0.121, though the difference is modest relative to cross-validation variability. Latent semantic analysis of TF-IDF captures topic-level structure that is sufficient for survival ranking. The contextual representations in MentalBERT add precision for binary identification. The temporal holdout further discriminates: MentalBERT degrades more (10%) than TF-IDF (6%), suggesting contextual representations are more sensitive to temporal distribution shift in lexical patterns.

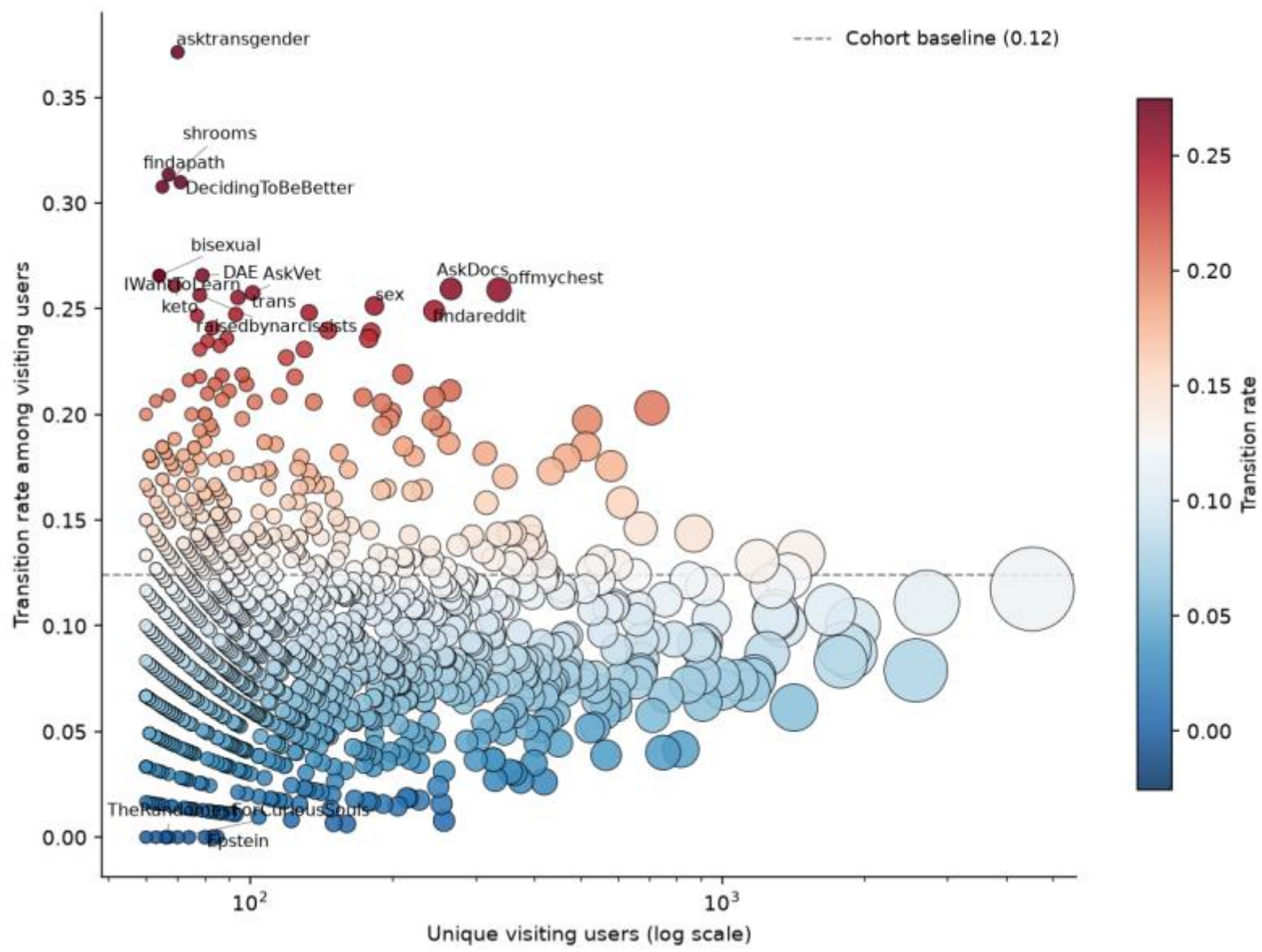

**Fig. 15.** Transition rate among visiting users versus number of unique visitors, by subreddit (control frame, subreddits with ≥60 visitors); dashed line marks the cohort baseline.

Fig. 16 confirms this qualitatively: a t-distributed stochastic neighbor embedding (t-SNE) [29] projection of user-level MentalBERT embeddings shows no clear linear separation between eventual transitioners and censored users, indicating that the embedding space captures broad semantic content rather than a directly separable help-seeking signal.

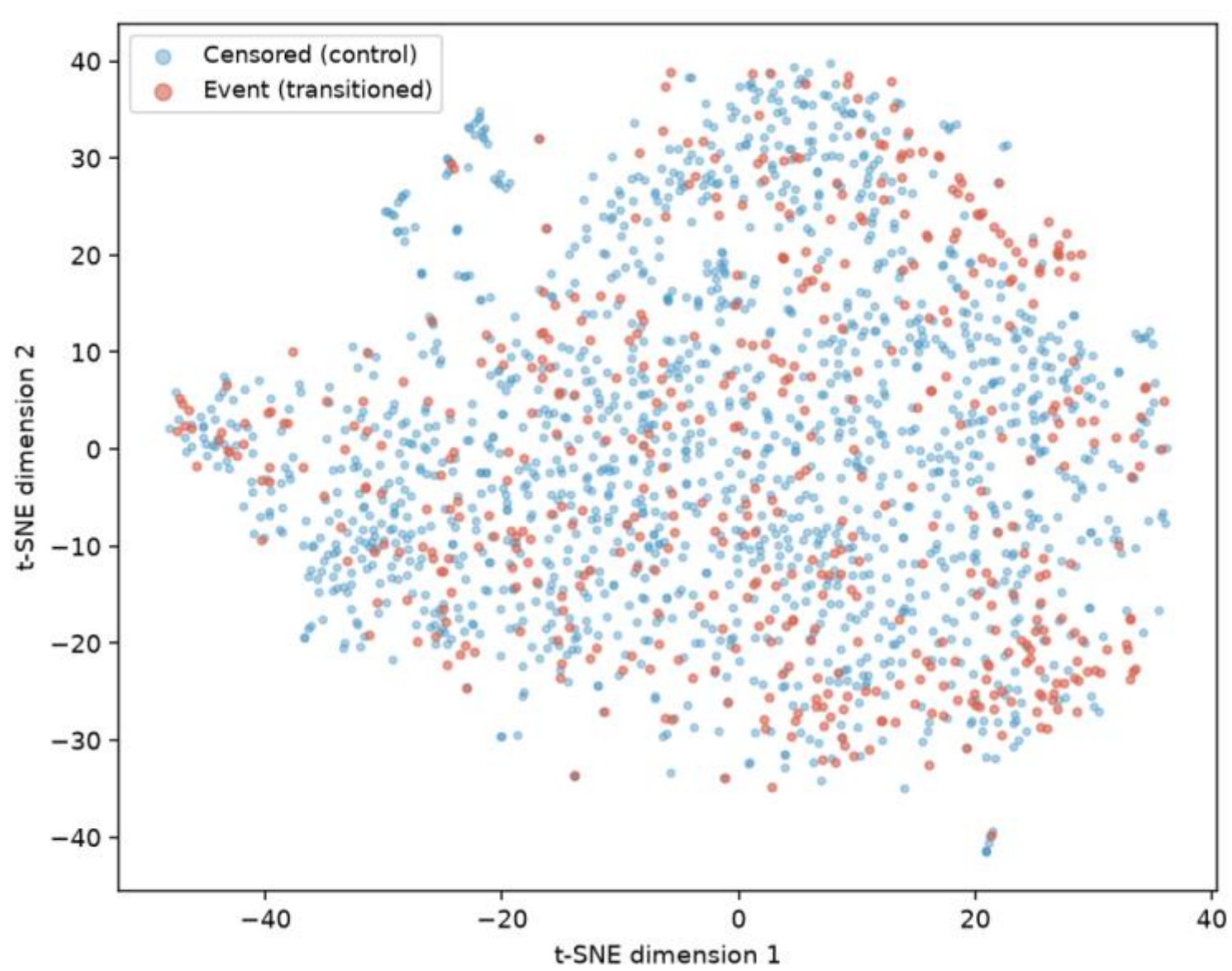

**Fig. 16.** t-SNE projection of user-level MentalBERT embeddings for a stratified sample of Arm 1 users (500 events, 1,500 censored), coloured by event status.

### 6.4. The two-arm decomposition

The dissociation between Arm 1 and Arm 2 results is theoretically important. Behavioural features are the dominant predictor of *who* (Arm 1, C-index 0.797/0.846), although combining them with text improves C-index further (0.823/0.856). Text features can predict *how soon* (Arm 2, C-index 0.698), contrary to behavioural features. This suggests that structural behavioural signals capture population-level risk factors for help-seeking, while linguistic signals carry individual-level information about the timing and urgency of a specific transition. A future monitoring system, if deployed with appropriate ethical safeguards and clinical oversight, might fruitfully combine both: behavioural risk stratification to identify high-priority users, and text-based imminence estimation to prioritise among them.

### 6.5. Relation to prior work

Our C-index of 0.763 (TF-IDF, Arm 1) is not directly comparable to prior work [4,5,8], since most of it frames the problem as classification on users already known to be clinical, not as survival analysis with proper censoring. Liu et al. [19] are a rare exception, fitting a Cox model to Reddit posts preceding transitions to r/SuicideWatch, but reporting a near-chance C-index of 0.5123. Their cohort has no genuine controls, and every user in it is already defined as someone who eventually transitions, so there is no real who-versus-when question left for their model to answer.

Our design keeps those two questions apart. On the analogous imminence task, using only clinical-frame users already known to transition (Arm 2), we reach a C-index of 0.698, and we think that gap comes mainly from this separation rather than from richer features. For the broader who-question, we report 0.823 under cross-validation and 0.856 on the holdout, achievable partly because our feature set is richer and the survival problem is correctly specified. Our contribution is therefore a methodological advance, not just a performance gain.

## 7. Limitations and ethics

### 7.1. Proxy labels, selection bias, and generalisability

First posting to a clinical subreddit is a proxy for help-seeking, not a clinical diagnosis; we make no claims about any individual's clinical state. Data are Reddit-only, English-only, and skewed toward male, younger, and more educated than the general population [30,31], a pattern consistent with broader concerns about non-representative samples in behavioural research [32]. Therefore, generalisability to other platforms or populations is not assumed.

### 7.2. Data completeness and temporal confounding

Right-censored users may have sought help elsewhere or beyond the cutoff date. Reddit's API also caps history at 1000 posts per listing, with unreliable coverage beyond roughly ten years; a widely reported constraint that motivated third-party archival efforts [33]. Together, these mean lead-time estimates (Arm 2, median 335 days) are lower bounds rather than exact durations. One behavioural feature, the recency gap, is confounded by construction: for cases, a short gap follows almost tautologically from the index-time definition rather than genuine signal. This is why we report robust results based on models excluding it (Section 5.7) and use these models wherever behavioural performance is discussed as a primary finding.

### 7.3. Proportional hazards assumption

Schoenfeld residual tests [34] show that three out of five behavioural features violate the proportional hazards assumption (log(N_posts) and mean upvote score: $p<0.001$; community diversity: $p=0.002$). This is consistent with the Random Survival Forest's advantage over Cox (C-index 0.941 vs. 0.797) in capturing non-linear, time-varying effects.

It is important to emphasize that these are research prototypes, not tools ready for clinical deployment without further validation. Table 2's coverage is deliberately scoped and non exhaustive: the Random Survival Forest covers TF-IDF, behavioural, and combined features, while the LSTM uses only MentalBERT's per-post embeddings, since TF-IDF and behavioural features are user-level aggregates with no per-post sequence to read.

### 7.4. Ethics

This research is observational and aggregate: no individual-level predictions or clinical decisions are made about real users. Content from r/SuicideWatch is used only as a label signal and never displayed; all posts are pseudonymous and results are reported in aggregate [35]. The study involves only computational analysis of public Reddit posts with no participant interaction. As the study involved no recruitment of, interaction with, or intervention on any individual, and analysed only publicly available pseudonymous posts, formal ethical review was not required under the University of Strathclyde's guidance for research without direct human interaction.

## 8. Future work

Several extensions could strengthen or help generalise the current results:

### 8.1. Broader data coverage

Both submissions and comments were used, but histories are limited by Reddit's 1,000-item listing cap (Section 7.2); supplementing the API with archival sources would deepen histories for users with sparse activity. The framework is platform-agnostic, and testing it on X (formerly Twitter), Facebook, or longitudinal clinical cohorts would show whether the behavioural-dominance finding can generalise.

### 8.2. Clinical readiness

Calibration curves (Section 5.9) overestimate event probabilities at the 12% base rate; isotonic regression [36] or Platt scaling [37], ideally with temporal recalibration, are needed before any risk score is interpreted [38]. Training directly for fixed-horizon classification (e.g., 90/180/365-day transition probability), rather than deriving it post-hoc from the survival curve, could sharpen near-term estimates.

### 8.3. Modelling extensions

Current behavioural features are purely aggregate; trajectory features (posting-rate slope, engagement trend, community-diversity growth) could capture dynamic change that static aggregates miss, particularly for Arm 2. A Transformer-based survival model [39] with interpretable attention weights could extend the LSTM's demonstrated value of post ordering, offering post-level attribution of which posts most predict transition.

## 9. Conclusion

We have presented a survival analysis framework for predicting mental health help-seeking behaviour on Reddit that addresses the methodological limitations of prior work: a two-arm cohort design with genuine censoring, temporal leakage prevention through user-grouped cross-validation, and appropriate survival metrics.

Our principal findings are:

1. Behavioural features (such as activity patterns, community diversity, and engagement) outperform text features for survival ranking in a population-level model (C-index 0.797 behavioural, 0.823 combined, both robust), with the advantage strengthening under a four-year temporal holdout (0.846 and 0.856 respectively).
2. MentalBERT and TF-IDF are complementary: domain-adapted embeddings improve binary identification (PR-AUC 0.277 vs. 0.244), while TF-IDF is a marginally better survival ranker (C-index 0.763 vs. 0.758).
3. Text content predicts imminence of transition (Arm 2 C-index 0.698), while behavioural patterns do not. This dissociation suggests different feature families answer different clinical questions.

This work contributes a replicable survival framing for social media mental health research and demonstrates that the *structural* aspects of how people use online communities encode predictive information about their future help-seeking behaviour that goes beyond what they say. This carries a practical implication for how outreach or screening tools might be designed. Methods based on language alone may correctly identify imminent risk among people already known to be struggling, but they can miss the earlier, broader signal carried by simple behavioural patterns such as posting frequency and community engagement, which point to who is heading toward a transition well before their language ever changes.

## Acknowledgments

Data collection used the PRAW Reddit API under read-only access.

## Data availability

Code for all stages (cohort construction, feature extraction, model training, and evaluation) is available from the corresponding author upon reasonable request. The raw Reddit post data cannot be redistributed under Reddit's API terms of service; the collection protocol is fully described in Section 3.1 and is sufficient to replicate the cohort.

## Funding

This research received no external funding.

## Declaration of competing interests

The authors declare that they have no known competing financial interests or personal relationships that could have appeared to influence the work reported in this paper.

## Ethics statement

This study used only publicly available, pseudonymous Reddit posts, collected read-only through Reddit's official API. There was no interaction with, recruitment of, or intervention on any individual, and no attempt at re-identification. No posts are quoted; all results are reported in aggregate, and content from crisis-support communities was used solely as an outcome signal. In line with the University of Strathclyde's guidance for research without direct human interaction, formal ethical review was not required.

## CRediT Authorship Contribution Statement

**Nisar Ahmed:** Conceptualisation, Data curation, Formal analysis, Methodology, Software, Writing – original draft. **Ammara Kazmi:** Conceptualisation, Data curation, Software. **Irtaza Bashir Raja:** Validation, Writing – review & editing. **Didier Devaurs:** Supervision, Writing – review & editing.

## Declaration of generative AI and AI-assisted technologies in the manuscript preparation process

During the preparation of this work the authors used Claude (Anthropic) in order to assist with drafting and editorial revision of the manuscript text. After using this tool, the authors reviewed and edited the content as needed and take full responsibility for the content of the published article.